%% file: collas2022_conference.tex
\documentclass{article} 
\usepackage{collas2022_conference,times}

\input{math_commands.tex}

\usepackage{hyperref}
\hypersetup{
    colorlinks=true,
    linkcolor=red,
    filecolor=magenta,      
    urlcolor=blue,
    citecolor=purple,
    pdftitle={Overleaf Example},
    pdfpagemode=FullScreen,
    }

\usepackage{url}

\usepackage{graphicx}
\usepackage{cleveref} 
\usepackage{wrapfig,lipsum,booktabs}
\usepackage{subcaption}
\usepackage{footmisc} 

\usepackage{listings}
\usepackage{color}
\usepackage{xcolor}

\definecolor{dkgreen}{rgb}{0,0.6,0}
\definecolor{gray}{rgb}{0.5,0.5,0.5}
\definecolor{mauve}{rgb}{0.58,0,0.82}

\newif\ifshowcomments

\showcommentsfalse

\definecolor{CommentBlue}{rgb}{0,0,0.7}

\newcommand{\rebuttal}[1] {{\ifshowcomments\color{CommentBlue} {#1}\else#1\fi}}
\newcommand{\revision}[1] {{\ifshowcomments\color{CommentBlue} {#1}\else#1\fi}}

\title{MO2: Model-Based Offline Options}

\author{
\hspace{0.23\textwidth} Sasha Salter\thanks{Work done while at DeepMind (sasha.salter@hotmail.com).}, Markus Wulfmeier, Dhruva Tirumala,
\\
\hspace{0.19\textwidth} \textbf{Nicolas Heess, Martin Riedmiller, Raia Hadsell, Dushyant Rao}   
\\
\hspace{0.35\textwidth} DeepMind, London, UK
}

\collasfinalcopy 

\begin{document}

\maketitle
\vspace{-5mm}
\begin{abstract}
The ability to discover useful behaviours from past experience and transfer them to new tasks is considered a core component of natural embodied intelligence. Inspired by neuroscience, discovering behaviours that switch at bottleneck states have been long sought after for inducing plans of minimum \revision{description} length across tasks. Prior approaches have either only supported online\rebuttal{, on-policy,} bottleneck state discovery, limiting sample-efficiency, or discrete state-action domains, restricting applicability. To address this, we introduce Model-Based Offline Options (MO2), an offline hindsight framework supporting sample-efficient bottleneck option discovery over continuous state-action spaces. Once bottleneck options are learnt offline over source domains, they are transferred online to improve exploration and \rebuttal{value estimation} on the \revision{transfer} domain. Our experiments show that on complex long-horizon continuous control tasks with sparse, delayed rewards, MO2's properties are essential and lead to performance exceeding recent option learning methods. Additional ablations further demonstrate the impact on option predictability and credit assignment.

\end{abstract}

\input{sections/introduction}
\input{sections/preliminaries}
\input{sections/method}

\input{sections/experiments}
\input{sections/results}
\input{sections/related_works}
\input{sections/conclusion}

\bibliography{collas2022_conference}
\bibliographystyle{collas2022_conference}

\input{sections/appendix}

\end{document}

%% file: math_commands.tex
\usepackage{amsmath,amsfonts,bm}

\def\eqref#1{equation~\ref{#1}}

\def\1{\bm{1}}

\DeclareMathAlphabet{\mathsfit}{\encodingdefault}{\sfdefault}{m}{sl}
\SetMathAlphabet{\mathsfit}{bold}{\encodingdefault}{\sfdefault}{bx}{n}



%% file: sections/introduction.tex
\vspace{-3mm}
\section{Introduction}
\label{sec:intro}

While Reinforcement Learning (RL) algorithms have recently achieved impressive feats across a range of domains \citep{silver2017mastering, mnih2015human, lillicrap2015continuous}, they remain sample-inefficient \citep{abdolmaleki2018maximum, haarnoja2018soft} which makes it challenging applying them to robotics applications. Natural embodied intelligence across their lifetime discover and reuse skills at varying levels of behavioural and temporal abstraction to efficiently tackle new situations. For example, in manipulation domains, beneficial abstractions could include low-level instantaneous motor primitives as well as higher-level object manipulation strategies. Endowing RL agents with a similar ability could be vital for attaining comparable sample efficiency \citep{parisi2019continual}.

The options framework \citep{sutton1999between} presents an appealing approach for discovering and reusing such abstractions, with options representing temporally abstract skills. We are interested in methods supporting sample-efficient embodied transfer learning \citep{khetarpal2020towards}, where skills learnt across prior tasks are used to improve performance on downstream ones. Of particular interest is the recent \emph{Hindsight Off-Policy Options (HO2)} \citep{wulfmeier2020data} supporting off-policy learning over continuous state-action spaces (crucial for sample reuse in embodied applications), training all options across every experience in hindsight (further boosting sample-efficiency). The latter is achieved by computationally-efficiently marginalising across all possible option-segmented trajectories during policy improvement, similar to \citet{shiarlis2018taco}, leading to state-of-the-art results on transfer learning benchmarks. Whilst \citet{wulfmeier2020data} considered discovering options online, we are interested in offline discovery, further encouraging sample-efficiency (discovering abstractions without any additional environment interactions). 

Even though the options framework supports skill discovery, it does not constrain their behaviour. As such, vanilla approaches often lead to degenerate solutions \citep{harb2018waiting}, with options switching every timestep, maximising policy flexibility and return, at the expense of skill reuse across tasks. Therefore, constraining skill properties may be necessary. Many traditional methods \citep{mcgovern2001automatic,csimcsek2004using, solway2014optimal, harutyunyan2019termination, ramesh2019successor} sought out \emph{bottleneck options} that terminate at bottleneck states. Bottleneck states are defined as the most frequently visited states when considering the shortest distance between any two states in an MDP \citep{solway2014optimal}, and are considered beneficial for planning by inducing plans of minimum description length across a set of tasks \citep{harutyunyan2019termination, solway2014optimal}. Intuitively, bottleneck states connect different parts of an environment, and are hence
visited more often by successful trajectories \citep{harutyunyan2019termination, mcgovern2001automatic, stolle2002learning}.  Unfortunately, these methods do not support off-policy hindsight learning, necessary for sample-efficiency, nor continuous state-action spaces, crucial in robotics.

To address this, we present \textit{Model-Based Offline Options} (MO2), an offline hindsight \emph{bottleneck options} framework supporting continuous \revision{(and discrete)} state-action spaces, that discovers skills suited for planning (in the form of \rebuttal{value estimation}) and acting across modular tasks. We refer to modular tasks as a family of MDPs whose optimal policies are obtained by recomposing \rebuttal{(a subset of)} shared temporal behaviours. \rebuttal{For example, navigation tasks with self-driving cars share a set of common temporal abstractions: traversal between junctions. While this modular constraint appears restrictive, options can collapse to the original action-space, transitioning per timestep if necessary.} MO2 discovers \emph{bottleneck options} by extending HO2 to the offline context, training options to maximise: 1) the log-likelihood of the offline behaviours; 2) a \textit{predictability objective} for option termination states across the episode, computed as the cumulative 1-step option-transition log-likelihood. Maximising the 1-step option-transition log-likelihood encourages low-entropic, predictable terminations. The cumulative equivalent (across all transitions), encourages minimal switching only at bottleneck states where behaviours diverge. Once options are learnt offline, we freeze them and perform online transfer learning over the option-induced semi-MDP, learning and acting over the temporally abstract skill space. We compare MO2 against state-of-the-art baselines on complex continuous control domains \citep{fu2020d4rl} and perform an extensive ablation study. We demonstrate that MO2's options are bottleneck aligned (see \Cref{fig:mo2_model}) and improve acting (and exploring), \rebuttal{value estimation}, and learning of a jumpy option-transition model (defined in \Cref{sec:mo2_model_sec}). On the challenging, sparse, long-horizon, AntMaze domain, MO2 drastically outperforms all baselines.

%% file: sections/preliminaries.tex
\vspace{-2mm}
\section{Preliminaries}
\vspace{-2mm}
\revision{
Our goal is to leverage temporally abstract skills extracted from large, unstructured and multi-task datasets (\emph{source domains}) to accelerate the learning of new tasks (\emph{transfer domain}). For skills to be beneficial for learning, they should be suited for planning and acting over them \citep{solway2014optimal}. We propose an offline \emph{bottleneck options} approach discovering skills with these properties, that supports continuous state-action spaces, for efficient skill transfer from large datasets. We decompose the problem of skill transfer into two sub-problems: 1) the extraction of skills suited for planning and acting from offline data, and 2) learning of transfer tasks over the temporally abstracted skill space.
}
\vspace{-2mm}
\subsection{Problem Formulation}
\vspace{-1mm}
Our work considers reinforcement learning in Markov Decision Processes (MDPs), defined by $M = (S, A, R, p, p^{0}, \gamma)$. $S$, $A$, $R$ denote state, action and reward spaces. $p(s'|s, a ):S\times S \times A \rightarrow \mathbb{R}_{\geq0}$ represents the dynamics model. $p^0(s): S \rightarrow  \mathbb{R}_{\geq0}$ represents the initial state distribution. We denote the history of states $s \in S$ and actions $a \in A$ up to timestep $t$ as $h_{t} = (s_{0}, a_{0}, s_{1}, a_{1}, \dots, s_t)$. In this work, we consider option-augmented policies, taking the \textit{call-and-return} formulation \citep{sutton1999between}, defining options as triple $(I (s_t, o_t), \pi^L(a_t|s_t, o_t), \beta(s_t, o_t))$. $I, \beta$ represent an option's initial and termination conditions and $\pi^L$ denotes its behaviour. Following \citet{Bacon2017, Zhang2019DAC}, we define $I = 1 \forall s_t \in S$ and sample a new option $o_t$ from $\pi^C(o_t| s_t)$ (the option-level controller) only if the previous one has terminated. $b(a|h)$ denotes the behaviour policy that collects experiences \revision{saved into the offline dataset $D$ in the form $(s_t, a_t)$. We leverage the options framework, trained on a maximum-likelihood behavioural cloning objective, to discover a set of task-agnostic, shared skills across tasks able to reconstruct offline behaviours. During transfer, we freeze these options, learning and acting over them to accelerate online RL on the new tasks. As such, we make the assumption that the offline datasets (\emph{source domains}) are multi-task and diverse enough that they exhibit all the temporal abstractions (skills) necessary for transfer. We refer to this as the \emph{modularity} assumption, that skills from \emph{source domains} suffice and can be recomposed to obtain optimal \emph{transfer domain} performance. In \Cref{sec:online_transfer_learning}, we discuss this assumption and how to relax it. 
}

\begin{wrapfigure}{r}{0.55\textwidth}
\vspace{-16mm}
  \begin{center}
    \includegraphics[trim={0 2mm 0 0},clip,width=0.53\textwidth]{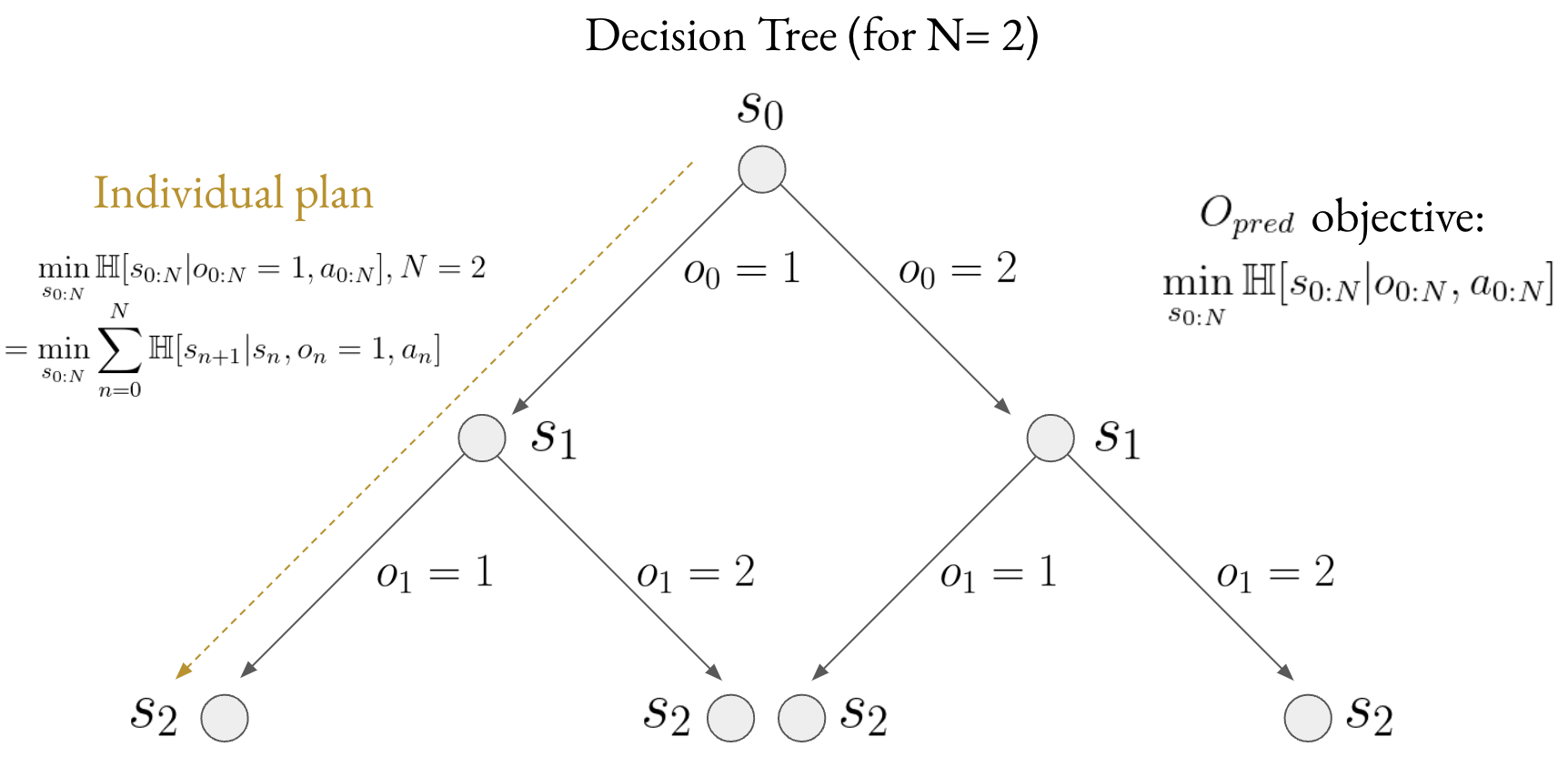}
  \end{center}
  \vspace{-3mm}
  \caption{\rebuttal{\emph{Option Decision Tree (depth = 2)}: MO2's $O_{pred}$ objective encourages low-entropic, compressed, option-level plans (over $s_n$) that reconstruct offline behaviours with high confidence. See \Cref{sec:offline_bottlenecks} for more details.}}
  \vspace{-5mm}
  \label{fig:decision_tree}
\end{wrapfigure}
\vspace{-2mm}
\subsection{HO2: Hindsight Off-Policy Options}
\label{sec:ho2}
\vspace{-1mm}
We build on Hindsight Off-Policy Options (HO2) \citep{wulfmeier2020data} as a highly sample-efficient, \revision{performant,} off-policy actor-critic \revision{options} algorithm \rebuttal{that builds on \citet{Smith2019}}. HO2 achieves higher sample efficiency than alternate options frameworks for two reasons: 1) HO2 trains all options in hindsight across all experiences within a trajectory, not just the current timestep. This is achieved by back-propagating gradients through HO2's dynamic programming inference procedure (discussed below), training all options end-to-end. 2) It uses Maximum-A-Posteriori Policy Optimisation (MPO) \citep{abdolmaleki2018maximum}, a highly performant, off-policy framework with monotonic improvement guarantees.

To train all options across all experiences, HO2 unrolls the graphical model for option-based policies and calculates the joint $\pi(a_t, o_t|h_t)$, marginalising across all combinations of option sequences $o_{0:t-1}$. To avoid a combinatoral explosion, HO2 uses the following recursive relation, similar to the one introduced in \citet{shiarlis2018taco}:
\begin{equation}
p(o_t|s_t, o_{t-1}) = \begin{cases}
1 - \beta(s_t, o_{t-1})(1 - \pi^C(o_t|s_t)) &\text{if $o_t=o_{t-1}$}\\
\beta(s_t, o_{t-1}) \pi^C(o_t|s_t) &\text{otherwise}
\end{cases}
\label{eq:recursive_calc}
\end{equation}
\vspace{-3mm}
\begin{equation}
\Tilde{\pi}^H(o_t|h_t) = \sum_{o_{t-1}=1}^{M} [p(o_t|s_t, o_{t-1})\pi^H(o_{t-1}|h_{t-1})\pi^L(a_{t-1}|s_{t-1}, o_{t-1})]
\label{eq:marg_calc}
\end{equation}
The distribution is normalized per timestep with $\pi^H(o_t|h_t) = \Tilde{\pi}^H(o_t|h_t) / \sum_{o'_t=1}^{M} \Tilde{\pi}^H(o'_t|h_t)$. \rebuttal{$\pi^H$ denotes option probabilities given histories and is not the same as $\pi^C$, the option controller for any given state.} \revision{Intuitively, option probabilities given history $\pi^H(o_{t}|h_{t})$ are not only dependent on the current state $s_t$ and option controller $\pi^C(o_t|s_t)$, but additionally the previous option (according to $\pi^H(o_{t-1}|h_{t-1})$) and whether it terminated (according to $\beta(s_t, o_{t-1})$). The exact relation is shown in \Cref{eq:recursive_calc,eq:marg_calc}. Using $\pi^H$ for optimisation, allows for all options up to timestep $t$ ($o_{0:t}$), not just the current option ($o_t$), to be optimised jointly on a control objective at time $t$ (e.g. for behavioural cloning $\log(\pi(a_t|h_t))$). As such, all options are trained across all experiences in the episode in hindsight, improving performance \citep{wulfmeier2020data}}. Building on option probabilities we obtain the joint\revision{:
\begin{equation}
 \pi(a_t,o_t|h_t) = \pi^L(a_t|s_t,o_t)\pi^H(o_t|h_t)   
\end{equation}
While \citet{wulfmeier2020data} uses the joint to effectively train $\pi^L,\pi^C,\beta$ on an online RL objective, we instead use an offline maximum-likelihood behavioural cloning objective, $O_{bc}(\pi^L,\pi^C,\beta) = \mathbb{E}_{a_t, h_t \sim D, o \sim \pi^H(\cdot|h_t)}[\log(\pi(a_t,o|h_t))]$ (omitting $t$ subscript from $o$ to emphasise it is not sampled from $D$), to distil skills from the offline dataset.}

%% file: sections/method.tex
\vspace{-2mm}
\section{MO2: Model-Based Offline Options}
\label{sec:mo2_model_sec}

\begin{figure}[h]
    \centering
    \includegraphics[width=0.9\textwidth]{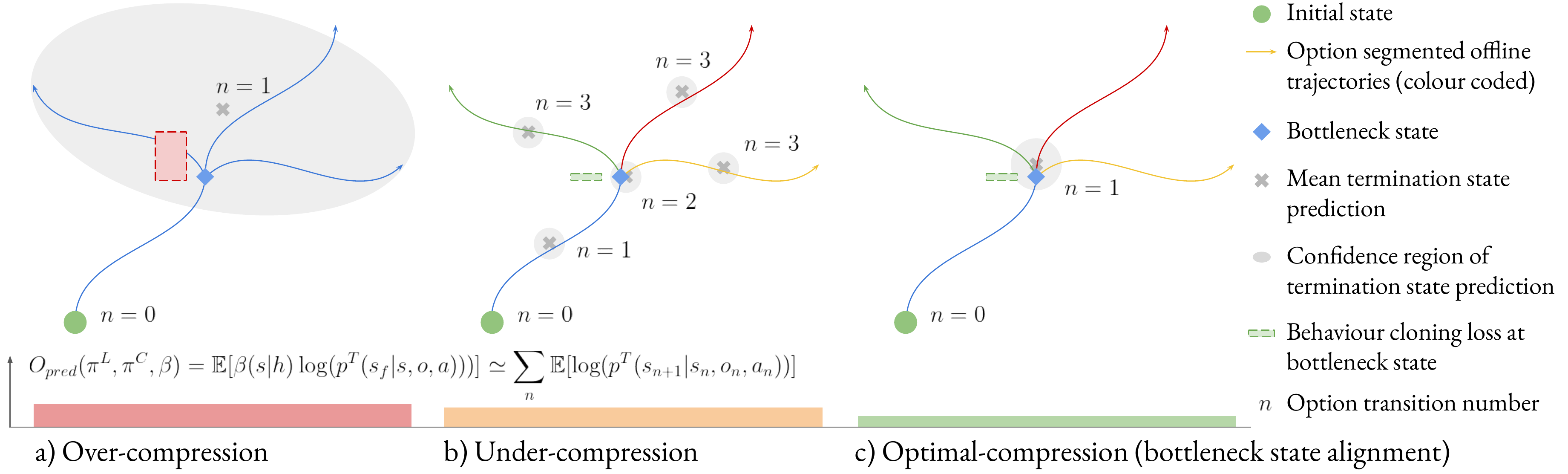}
    \vspace{-3mm}
    \caption{\emph{Intuition behind MO2}. MO2 trains an option policy and option-transition model, $\tilde{p}^T(s_f|s,o,a)$, predicting terminal state $s_f$ of option $o$ given state $s$ and action $a$. MO2's \emph{predictability} objective, $O_{pred}$, represents the predictability of option-transitions, $p^T(s_{n+1}|s_n, o_n, a_n)$, across the episode, with $n$ the option-transition number, $s_{n+1}$/$s_n$ the $n^{th}$ options termination/initiation states. Individual log-likelihoods ($n$) encourage predictable, low-entropic transitions, minimising any termination confidence region (individual ellipse area), ruling out \emph{a) over-compression}. Given positive transition entropies, the cumulative log-likelihoods (across $n$s) lead to transitions only when necessary, to reduce cumulative entropies (intuitively similar to the total area across ellipses), ruling out \emph{b) under-compression}. MO2's \emph{behavioural cloning} objective, $O_{bc}$, further encourages terminations where multi-modal behaviours exist. Combined, MO2 leads to \emph{c) optimal-compression (bottleneck state alignment)}, transitioning only at \emph{bottlenecks states}.}
    \label{fig:mo2_model}
    \vspace{-2mm}
\end{figure}

While HO2 provides a sample-efficient framework for discovering options, no constraints are placed on their behaviour. \revision{As such, it is unclear whether HO2's abstractions are suited for \emph{compressed} decision making across tasks, arguably the key reason of temporal abstractions as tools for computation- and sample- efficient learning. Taking inspiration from humans as resource constrained learners, \citet{solway2014optimal} demonstrate the favouring of abstractions suited for long horizon reasoning (specifically \emph{planning}) and \emph{acting} across tasks. Particularly suited abstractions are those that transition between environment \emph{bottleneck states},} concentrated regions of the state-space that are commonly visited \revision{between tasks} when traversing distinct and diverse parts of the environment (such as doors \revision{or junctions, for various navigation tasks, in multi-room}\revision{ or maze domains, respectively}) \citep{solway2014optimal, harutyunyan2019termination}.
\revision{As such, bottlenecks give rise to a \emph{compressed} decision problem, reasoning only where necessary, over a few key decision making states (predictable due to their concentrated nature) where behaviours diverge.}
\revision{\emph{Bottleneck options} methods discover such abstractions, suited for long horizon option-level \emph{planning} (due to the high \emph{predictability} and \emph{compressed} nature of option-transitions) and \emph{acting} across tasks.} Unfortunately, existing methods \citep{mcgovern2001automatic,stolle2002learning,solway2014optimal,harutyunyan2019termination} do not support offline hindsight learning, necessary for sample-efficiency, nor continuous state-action spaces, crucial in robotics. \revision{We propose a method without these shortcomings, \emph{compressing} diverse multi-task offline behaviours into \emph{bottleneck options}.}

\subsection{Discovering Bottleneck Options Offline}
\label{sec:offline_bottlenecks}

We introduce Model-Based Offline Options (MO2), an offline hindsight \emph{bottleneck options} method supporting continuous state-action spaces. \revision{MO2 distils offline behaviours into a set of skills by training on the maximum-likelihood behavioural cloning objective $O_{bc}$ introduced in \Cref{sec:ho2}, using HO2's efficient calculation of the joint $\pi(a,o|h)$ to optimise all options in hindsight. However, in addition to training on $O_{bc}$, MO2 extends HO2 with an additionally learnt option-level transition model, and a \textit{predictability}
objective $O_{pred}$ encouraging option-level transitions across the episode to be predictable, compressed and beneficial for planning, thereby discovering \emph{bottleneck options}}:  

\vspace{-6mm}
\begin{equation}
    \rebuttal{O_{pred} = \sum_{n=0}^{N}\mathbb{E}_{s_n, o_n, a_n, s_{n + 1} \sim \pi}[\log(p^T(s_{n+1}|s_n, o_n, a_n))] = -\sum_{n=0}^{N} \mathbb{H}_{\pi}[s_{n + 1} | s_n, o_n, a_n] = -\mathbb{H}_{\pi}[s_{0:N}|o_{0:N}, a_{0:N}]}
    \label{eq:o_pred}
\end{equation}
\vspace{-6mm}

Here, $n$ is the option-transition number within the episode, \rebuttal{$N$ the maximum,} $s_{n+1}$ and $s_n$ are the $n^{th}$ option's termination and initiation states respectively, and $p^T(s_f|s, o, a)$ is the option-transition distribution over the terminal state $s_f$ of an option $o$, given state $s$ and action $a$. \rebuttal{Note that $n \neq t$ as options act at a different timescale. Shown in \Cref{eq:o_pred} (for proof see \Cref{sec:all_proofs}), encouraging predictable option-transitions is equivalent to encouraging individual conditional entropies to be low $\mathbb{H}_{\pi}[s_{n + 1} | s_n, o_n, a_n]$, ruling out the \emph{over-compression} of offline behaviours (see \Cref{fig:mo2_model} for an explanation). Maximising $O_{pred}$ is additionally equivalent to minimising the joint conditional entropy of option transition states $s_{0:N}$ given options $o_{0:N}$ and first option-executed actions $a_{0:N}$ from policy $\pi$, $\mathbb{H}_{\pi}[s_{0:N}|o_{0:N}, a_{0:N}]$ (see \Cref{sec:all_proofs} for proof and \Cref{fig:decision_tree} for a visual depiction). We subscript entropies by $\pi$ to emphasise they are dependent on the policy. If individual transition entropies are positive (see \Cref{eq:MO2_loss_detailed} for how this is achieved), then $O_{pred}$ encourages minimal option-transitions able to reconstruct offline behaviours using options, as to minimise $N$ and the accumulation of positive entropies. As such, $O_{pred}$ rules out the \emph{under-compression} and leads to the \emph{optimal-compression (bottleneck state alignment)} scenario in \Cref{fig:mo2_model}. For a more detailed explanation as to why this objective discovers \emph{bottleneck options} we refer the reader to \Cref{sec:why_bottlenecks}.}

\Cref{eq:o_pred} requires evaluating the agent $\pi$ in the environment to obtain $s_{n+1}, s_{n}$. As such, this objective does not support offline learning, as desired. Additionally, inline with HO2, we wish to train all options across all experiences in hindsight, as to maximise sample-efficiency. Therefore, we present an alternate form for $O_{pred}$ supporting offline hindsight learning, unrolling the graphical model for option-based policies and training across a distribution of option-transitions given histories (by efficiently marginalising over option transitions $o_{0:t-1}$ using \Cref{eq:recursive_calc,eq:marg_calc}):
\begin{equation}
    O_{pred} = \mathbb{E}_{\substack{s_t, h_t \sim D \\ o \sim \pi^H(\cdot|h_t), a \sim \pi^L(\cdot|s_t, o) \\ s_f \sim p^T(\cdot|s_t, o, a)}} [\beta(s_t|h_t)\log(p^T(s_f|s_t,o,a))]
    \label{eq:o_pred_offline}
\end{equation}
With $\beta(s_t|h_t)=\sum_{o_{t-1}=1}^{M}\beta(s_t, o_{t-1})\pi^H(o_{t-1}|h_{t})$, representing the probability that state $s_t$ is terminal, for the previous option $o_{t-1}$, and initial for the subsequent option $o_t$, given history $h_t$. $\pi^H(o_{t-1}|h_{t})=\pi^H(o_{t-1}|h_{t-1})\pi^L(a_{t-1}|o_{t-1}, s_{t-1})/\pi(a_{t-1}|h_{t-1})$ represents the distribution over $o_{t-1}$ given history $h_t$. \rebuttal{$D$ represents the offline data created by behaviour policy $b(a|h)$. For $O_{pred}$, options are sampled from $\pi^H(o_t|h_t)$ as $D$ does not contain options.}\revision{ In \Cref{eq:o_pred_offline}, we do not subscript $o$ and $a$ with $t$ to emphasise that these samples are not from $D$.} This form of $O_{pred}$ uses $\beta$ as a weighting for how probable $s_t$ is a starting state ($s_n$) of an option. The expectation is now taken over offline trajectories, with $\beta$ proportionally representing the summation over option-transitions, akin to the original $O_{pred}$. We use $p^T$ to sample terminal states $s_f$ ($s_{n+1}$) in \Cref{eq:o_pred_offline}. In practice, we do not have access to $p^T$ and instead learn it offline (\Cref{sec:transition_model_learning}).
\rebuttal{See \Cref{sec:why_both_o_pred_terms_equivalent} for details under which conditions \Cref{eq:o_pred,eq:o_pred_offline} are equivalent from an optimisation perspective.} Our overall objective combines $O_{pred}$ with \revision{$O_{bc}$}:
\begin{equation}
\underset{\pi^L, \pi^C, \beta}{\max}O_{mo2}(\pi^L, \pi^C, \beta) = \mathbb{E}_{\substack{s_t, a_t, h_t \sim D \\ o \sim \pi^H(\cdot|h_t), a \sim \pi^L(\cdot|s_t, o) \\ s_f \sim p^T(\cdot|s_t, o, a)}} [\beta(s_t|h_t)\log(p^T(s_f|s_t,o,a)) + \log(\pi(a_t, o|h_t))]
\label{eq:MO2_loss}
\end{equation}
While $p^T(s_f|s_t, o, a)$ encourages terminations predictable across all encountered states $s_t$, the additional weighting $\beta(s_t|h_t)$ focuses predictability only over the distribution of (rather than sampled) option initiation states given histories. In line with the intuition from \Cref{fig:mo2_model}, $O_{pred}$ and $O_{bc}$ together encourage offline hindsight \emph{bottleneck options} discovery over continuous state-action spaces.

\subsection{Learning the Option-Transition Model Offline}
\label{sec:transition_model_learning}

MO2 assumes access to the option-level transition distribution $p^T(s_f|s,o,a)$ in \Cref{eq:MO2_loss}. Given that we do not have access to such a model, we instead learn to approximate it, $\tilde{p
}^T(s_f|s,o,a)$, using a two-step self-supervision process. We first sample options \revision{$o \sim \pi^H(o_t|h_t)$} and then sample terminations according to $p^T(s_f| s_{t:T}, a_{t:T}, o)$, the termination distribution of the option given future states and actions. \revision{We do not subscript $o$ with $t$ to emphasise that these samples are not from $D$.} Specifically, we sample $s_f$ by sampling termination condition $\beta(s_{k}|o)$ over all future timesteps, $k \in [t, T]$, with $T$ representing episodic length. Each time the termination sample is true, the corresponding state is labelled terminal, $s_f = s_{k}$.
\begin{equation}
\underset{\tilde{p}^T}{\max}\;\;O_{tran}(\tilde{p}^T) = \mathbb{E}_{\substack{s_{0:T}, a_{0:T}, h_t \sim D \\ o \sim \pi^H(\cdot|h_t) \\ s_{f} \sim p^T(\cdot| s_{t:T}, a_{t:T}, o)}} [\log(\tilde{p}^T(s_{f}|s_t,o,a_t)))]
\label{eq:O2T_loss}
\end{equation}
As such, we learn an option-transition model that predicts the termination state distribution of an option for any state during its execution (not just the initial state). We note that this objective is biased, as the future states $s_{t:T}$ and actions $a_{t:T}$ used to sample $s_f$ are not from the option-policy $\pi$ over which we predict option-transitions, but the behaviour policy $b$ that created the offline behaviours. However, over time, as $\pi(s_{t:T}, a_{t:T}|h_t)$ aligns with $b(s_{t:T}, a_{t:T}|h_t)$, as encouraged by \revision{$O_{bc}$}, this bias tends to zero. We train $\tilde{p
}^T(s_f|s_t,o,a)$ using \Cref{eq:O2T_loss} while simultaneously using it to train \revision{$\pi^L, \pi^C, \beta$} with \Cref{eq:MO2_loss}. During transfer, the options aid acting, exploration, and \rebuttal{value estimation}.

\subsection{Online Transfer Learning}
\label{sec:online_transfer_learning}

Once options are learnt offline (over \emph{source domains}), there are several ways one can use them to improve online learning on a \emph{transfer domain}. \revision{We make the \emph{modularity} assumption, that skills exhibited over \emph{source domains} suffice for optimal \emph{transfer domain} performance when recomposed.} As such, during transfer, MO2's options are frozen and we train a new high-level categorical controller, $\pi^C(o_t| s_t)$, to recompose them, using MPO (see \Cref{app:online_transfer}). \rebuttal{While the \emph{modular} constraint may appear restrictive, the increasingly diverse the \emph{source domains} are (arguably as desired during lifelong learning \citep{khetarpal2020towards}), the increasingly probable that the optimal \emph{transfer} policy can be obtained by recomposing a subset of the \emph{source} abstractions. If this assumption does not hold, one could either fine-tune the existing options during \emph{transfer}, which would require tackling the catastrophic forgetting of skills \citep{kirkpatrick2017overcoming}, or train additional new options. We leave this as future work.} MO2's temporal abstractions afford two benefits: 1) temporally-correlated actions and exploration (over bottleneck states); 2) improved credit assignment and value estimation over the option induced semi-MDP (with a lower temporal granularity than the original MDP and occurring over concentrated regions of the state-space deemed beneficial for planning \citep{solway2014optimal}).

%% file: sections/experiments.tex
\vspace{-3mm}
\section{Experimental Setup}
\label{main_questions}
\vspace{-2mm}
In this paper, we explore how beneficial MO2's behavioural abstractions are for: 1) acting and exploration; 2) \rebuttal{value estimation}; 3) learning an option-transition model; 4) discovering compressed abstractions terminating at bottleneck states. We compare our results against two offline versions of HO2 trained solely to maximise $O_{bc}$ and not on $O_{pred}$. We call these baselines HO2 (offline) and HO2-lim (offline), a variant introduced in \citet{wulfmeier2020data} \revision{(see \Cref{app:offline_skill_discovery})} 
encouraging minimal option switching by penalising switches (yet does not encourage predictable terminations). We compare against these baselines to inspect how important the predictability of options are for all the these questions. Additionally, we compare with HO2 as we build on it as a state-of-the-art, sample-efficient, options framework supporting continuous state-action spaces. Finally, we compare against HO2 trained from scratch on the transfer task (HO2 scratch) to quantify the importance of skill transfer. See Appendix for full experimental details. 

\subsection{Semi-MDP vs MDP Ablations}
\vspace{-1mm}
To investigate the relative importance of MO2's options for both exploration and \rebuttal{value estimation}, we run two variants of each method during transfer: 1) \rebuttal{value estimation} occurs over the semi-MDP; 2) \rebuttal{value estimation} occurs over the MDP (akin to the original HO2 that we build on \citep{wulfmeier2020data}). By contrasting these experiments we can infer the importance of MO2's temporal abstractions, in relation to others, for credit assignment. Comparing against the HO2 baselines, we can evaluate the importance of MO2's options for exploration. For all methods with pre-trained options, we act at the option-level during transfer. See \Cref{app:alg_details} for further ablation information.
\vspace{-2mm}
\subsection{Domains}
\vspace{-2mm}
We evaluate on two D4RL suite domains \citep{fu2020d4rl}: Maze2D and AntMaze \revision{(see \Cref{fig:policy_rollouts,fig:d4rl_setup} and \Cref{app:environments} for details). The D4RL suite was designed to test the ability of learners to leverage large, diverse, unstructured, multi-task datasets (\emph{source domains}) to accelerate learning of downstream tasks (\emph{transfer domains}). Availability to such datasets are expected for lifelong learners \citep{khetarpal2020towards} and becoming increasingly abundant in robotics, with their wide-range deployment on streets \citep{caesar2020nuscenes}, offices \citep{mo2018adobeindoornav}, and warehouses \citep{dasari2019robonet,cabi2019scaling}, collecting diverse experiences. We evaluate on two navigation domains, with distinct embodiments (pointmass for Maze2D; quadruped for AntMaze),  as they are representative of many challenging real-world robotic settings, such as autonomous driving \citep{caesar2020nuscenes}. Navigation domains are particularly challenging due to their long-horizons and sparse rewards, making exploration and credit-assignment across extended timescales difficult. We test MO2's ability to discover temporal abstractions beneficial for \emph{transfer} in these settings. 

Specifically, the offline data (\emph{source domains}) corresponds to trajectories created by optimal, shortest route, goal reaching policies for randomly initiated agent and goal locations across episodes (see \Cref{fig:d4rl_setup}). The maze layout is not randomised. Between tasks, behaviours only diverge at corridor intersections, based on which corridor ordering leads to the shortest path to the goal. As such, these are the \emph{bottleneck states} of this setup, leading to plans of minimum description length across tasks \citep{solway2014optimal} when planning over them. The \emph{transfer domain} keeps the same maze layout but has an unknown, fixed, goal location. The agent is always spawned on the other side of the maze and must discover where the goal is, and the shortest path to it (by traversing a new ordering of corridor intersections, the \emph{bottleneck states}). The \emph{transfer domain} has sparse rewards (rewarded $+1$ only upon goal reaching) and long-horizons.We compare Maze2D and AntMaze as both have distinct state (2 vs 111) and action (2 vs 8) dimensionalities, and episodic-lengths (200 vs 900 for optimal policies), allowing us to inspect how well MO2 scales (specifically its predictability objective) across these dimensions and to increasingly hard-exploration domains. Additionally, as mentioned in \citet{fu2020d4rl}, AntMaze is introduced to mimic a more challenging real-world robotic navigation task.} 

%% file: sections/results.tex
\vspace{-6mm}
\section{Results}
\vspace{-2mm}
In this section, we perform a qualitative and quantitative analysis of the options belonging to MO2 and the baselines, and in the following sections answer the questions from \Cref{main_questions}. All tables in this section report mean $\pm$ one standard deviation, obtained across four random seeds. We additionally report the maximum and minimum performance, either theoretically or achieved across the learners and their lifetimes. As such, we contextualise the values in each table. It should be clear which performance we report.

\vspace{-2mm}
\subsection{Temporal Compression Of Offline Behaviours (Bottleneck Options)}
\begin{figure}[h]
 \centering
 \includegraphics[width=0.24\textwidth, totalheight=29mm, trim={27mm 27mm 27mm 27mm}, clip]{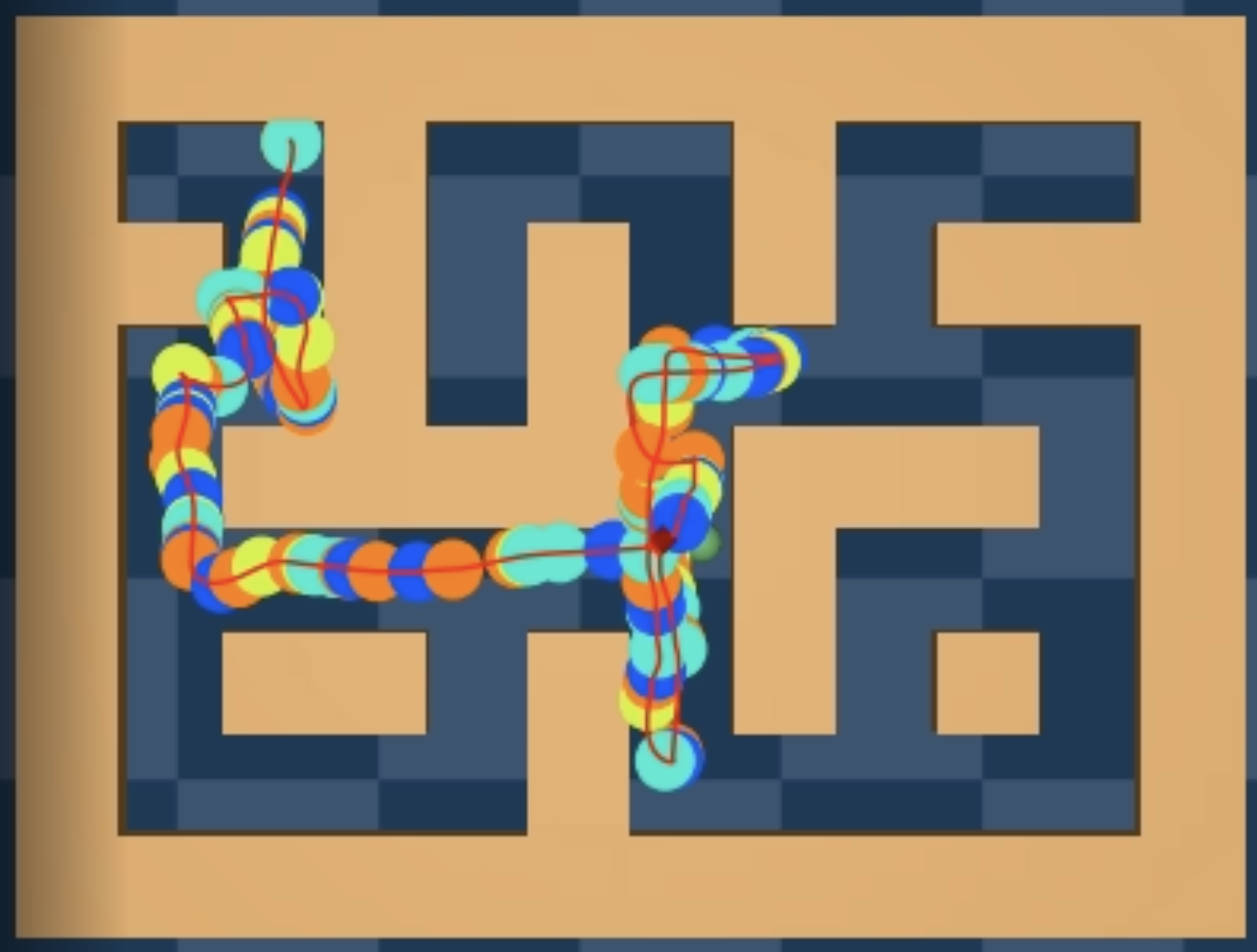}
 \includegraphics[width=0.24\textwidth, totalheight=29mm, trim={27mm 27mm 27mm 27mm}, clip]{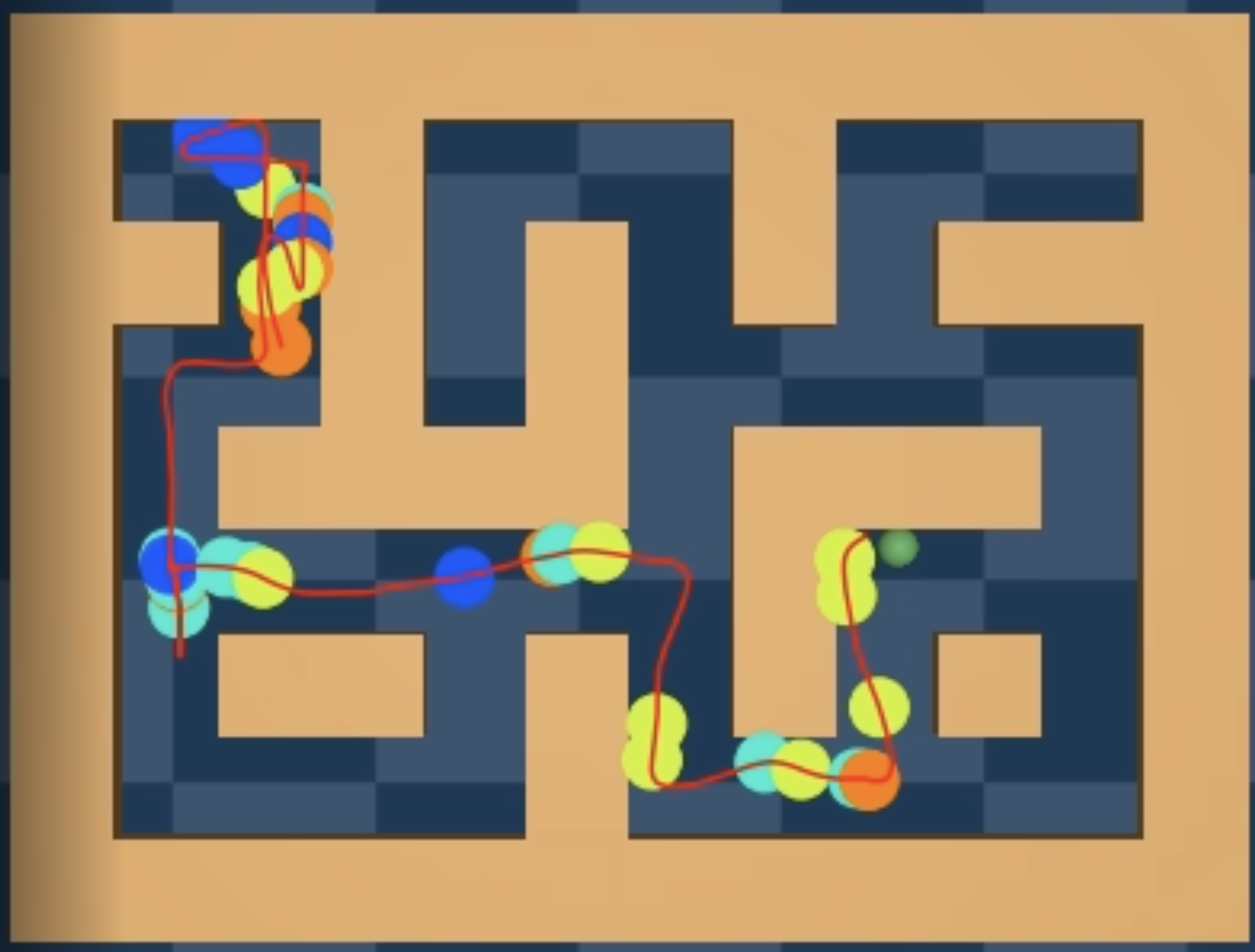}
 \includegraphics[width=0.24\textwidth]{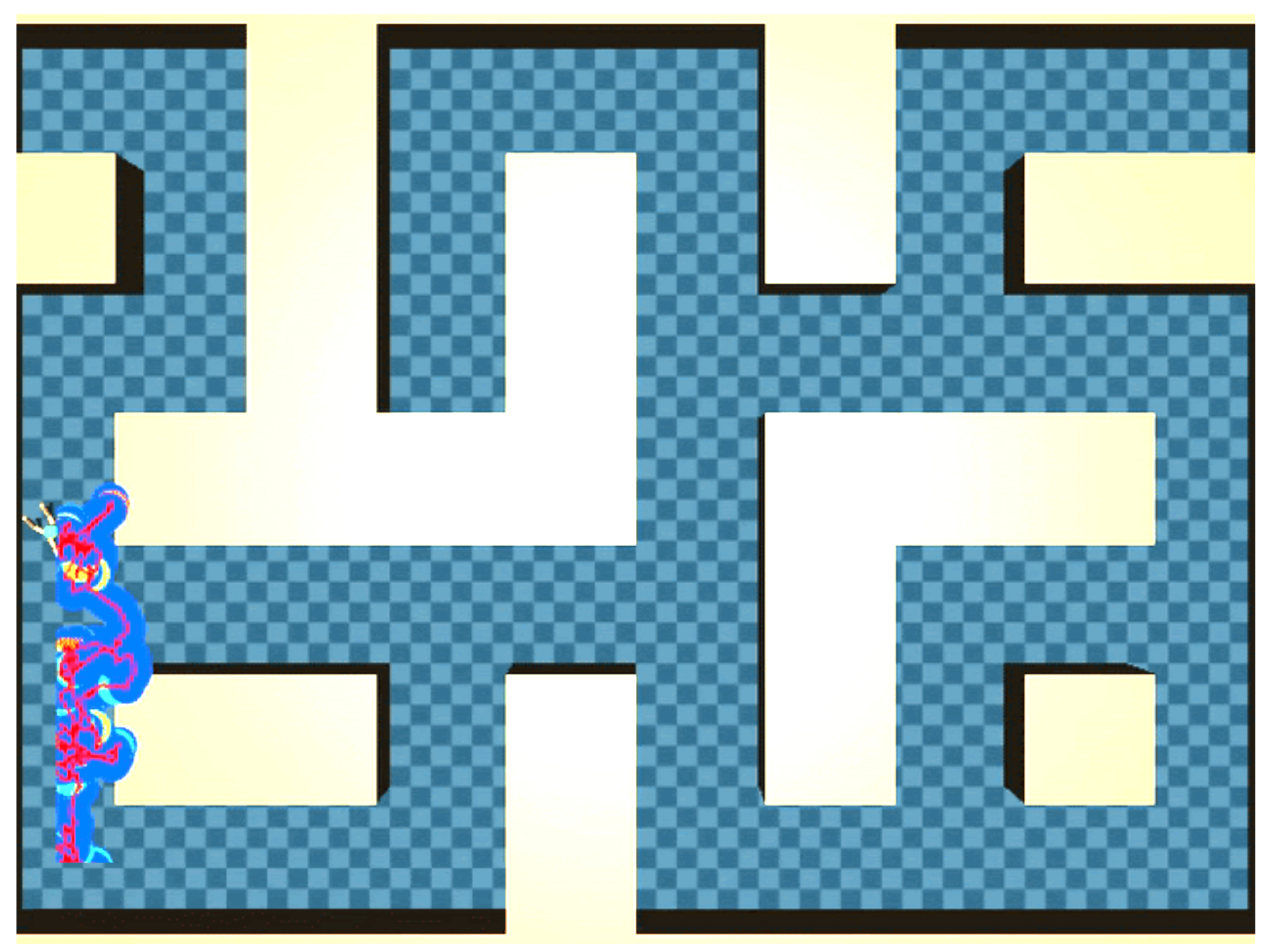}
 \includegraphics[width=0.24\textwidth]{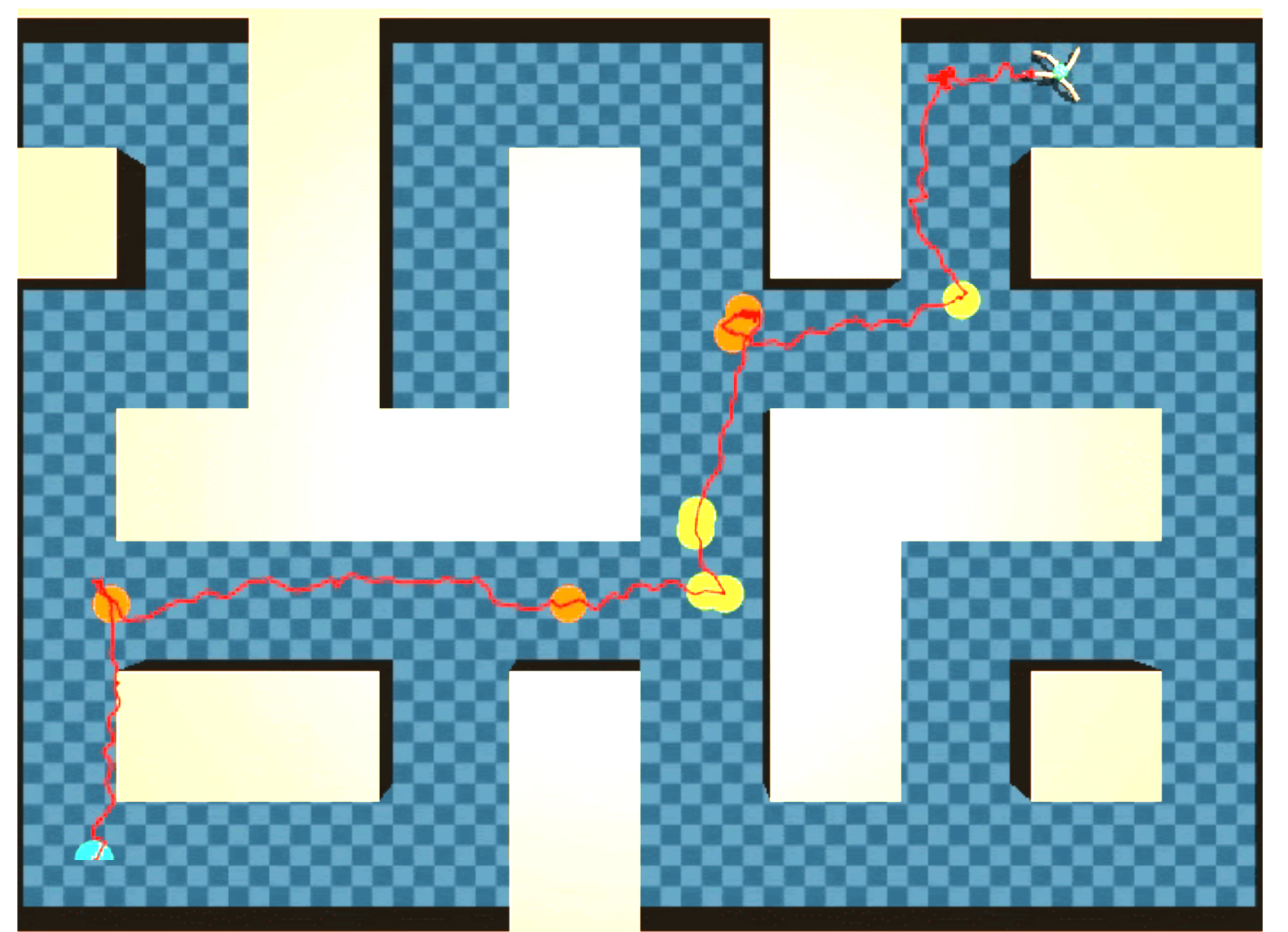}
 \caption{\emph{Policy Rollouts}. We plot policy rollouts across $1$ episode for Maze2D (left two plots) and AntMaze (right two plots), for HO2-lim (offline) (inner-left for each domain) and MO2 (inner-right for each domain). These rollouts are obtained early during training on the transfer domain. The centre of mass trajectory is plotted as a red line. The termination locations of options are denoted as coloured circles along the rollout, with the colour denoting the subsequent chosen option. MO2 leads to more temporally compressed options, switching less frequently, and primarily at environment-aligned, bottleneck states, corresponding with corridor intersections. As such, MO2 leads to temporally correlated behaviour, traversing the depths of the maze. In contrast, HO2-lim (offline) exhibits high frequency switching with shallower maze traversal for the longer horizon, higher dimensional, AntMaze domain.}
 \label{fig:policy_rollouts}
\end{figure}
\vspace{-2mm}
\begin{table}[h]
    \scriptsize
    \centering
    \caption{Temporal Compression Metrics}
    \begin{subtable}[h]{0.48\textwidth}
    \centering
      \caption{Average Switch Rate}
  \label{tab:average_switch_rate}
   \begin{tabular}{lll}
 \toprule
 \multicolumn{1}{c}{Environment} & \multicolumn{1}{c}{Maze2D} & AntMaze  \\
 \toprule
 HO2 (offline) & $0.48\pm0.04$ & $0.56\pm0.04$\\
 HO2-lim (offline) & $0.15\pm0.01$ & $0.36\pm0.01$\\
 MO2 & $\mathbf{0.03\pm0.00}$ & $\mathbf{0.01\pm0.00}$\\
 \toprule
 Max (min) & $1.00$ $(0.00)$ & $1.00$ $(0.00)$ \\
 \bottomrule
 \end{tabular}
  \end{subtable}
  \hfill
      \begin{subtable}[h]{0.48\textwidth}
      \centering
      \caption{Expected Behaviour Cloning Performance
  }
        \label{tab:bc_loss}
 \begin{tabular}{lll}
 \toprule
 \multicolumn{1}{c}{Environment} & \multicolumn{1}{c}{Maze2D} & AntMaze \\
 \toprule
 HO2 (offline) & $\mathbf{0.102\pm0.004}$ & $\mathbf{1.30\pm0.05}$\\
 HO2-lim (offline) & $0.082\pm0.005$ & $0.98\pm0.02$\\
 MO2 & $0.079\pm0.001$ & $\mathbf{1.27\pm0.04}$\\
 \toprule
 Max (min) & $0.102$ $(-5.000)$ & $1.30$ $(-50.00)$ \\
 \bottomrule
 \end{tabular}
  \end{subtable}
\end{table}

We analyse the properties of the options for each method. In \Cref{fig:policy_rollouts}, we visualise representative rollouts of the transfer agent, early during training, using the pre-trained, frozen, options. We plot rollouts for MO2 and HO2-lim (offline). We plot centre-of-mass rollouts for each domain, depicted as a red line, together with option termination locations, depicted as circles colour coded by the subsequent chosen option. 

For both domains, we observe that MO2's options primarily align with individual corridor traversal with terminations occurring at the intersection of corridors. MO2's options have therefore discovered the environment bottleneck states (corridor intersections) leading to switching of options at a low-entropic distribution of states that are highly visited and predictable, where behaviours naturally diverge, connecting distinct regions of the environment (individual corridors). This leads to high temporal compression of offline behaviours, with option switches occurring infrequently.

In contrast, HO2-lim (offline) exhibits far lower temporal compression, with option switching occurring more frequently, and at seemingly random locations, not aligning with bottleneck states. This suggests that penalising option switches, as achieved through HO2-lim (offline), is not enough to achieve effective temporal compression and bottleneck discovery. Comparing average switch rates (\revision{the inverse of expected option duration over offline trajectories}), seen in \Cref{tab:average_switch_rate}, additionally shows the compression disparity between each approach. Importantly, MO2's increased temporal compression barely influences controllability, as seen by comparing behaviour cloning $O_{bc}$ values in \Cref{tab:bc_loss} (over offline trajectories, the higher the score the better). In \Cref{tab:bc_loss}, max (min) scores refer to the range of values achieved across all methods during training. \rebuttal{We report mean $\pm$ one standard deviation, across four random seeds.}
\vspace{-3mm}
\subsection{Exploration}
\begin{figure}[h]
 \centering
 \includegraphics[width=0.75\textwidth]{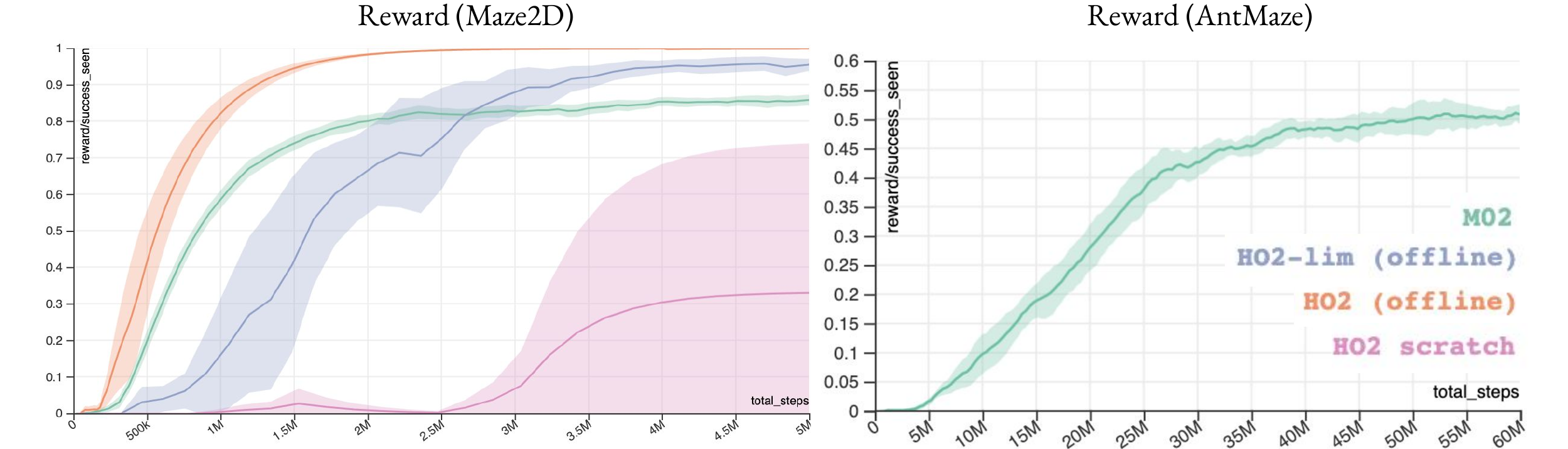}
 \vspace{-3mm}
 \caption{\emph{Return Curves}. We plot return learning curves for both transfer environments and all methods. Experiments were ran with 4 seeds, and we plot mean (solid line) and $1$ standard deviation (shaded region). Curves are periodically obtained by evaluating the current policy on the environment. Temporally compressed, bottleneck aligned, abstractions are essential for the sparse, long horizon, AntMaze, with MO2 the only method with non zero reward. For the less sparse, Maze2D, domain, MO2's abstractions are less necessary and hinder asymptotic performance.
 \vspace{-3mm}}
 \label{fig:mo2_return}
\end{figure}
\vspace{-0mm}

In \Cref{fig:mo2_return}, we plot transfer performance for each approach on both domains. For the sparser, longer horizon, higher-dimensional action-space AntMaze domain, we observe that MO2 is the only method able to attain any rewards and solve the task, demonstrating the benefits of MO2's temporally compressed bottleneck options for exploration. In contrast, for Maze2D, temporally compressed behaviours, achieved either by MO2 or HO2-lim (offline), do not yield benefits, demonstrating that for simpler exploration domains, temporal compression is not necessary for effective transfer, and can slightly hinder policy flexibility and asymptotic performance. Comparing initial policy rollouts in \Cref{fig:policy_rollouts}, we see that MO2's reduced switch rate leads to more directed, temporally correlated, exploration, with wider coverage of the maze for AntMaze. This is not the case for Maze2D, suggesting that for low-dimensional environments with short horizons, 
compression is not necessary for exploration.

\vspace{-3mm}
\subsection{\rebuttal{value estimation}}
\begin{figure}[h]
\vspace{-5mm}
 \centering
 \includegraphics[width=0.8\textwidth]{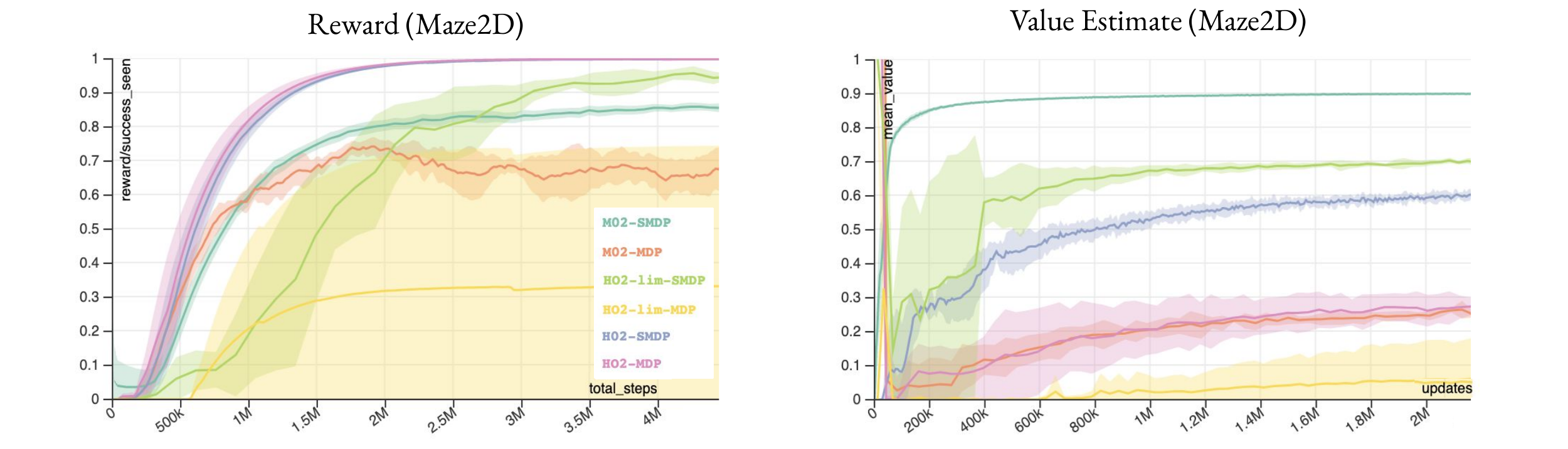} \\
 \vspace{-1mm}
 \includegraphics[width=0.8\textwidth]{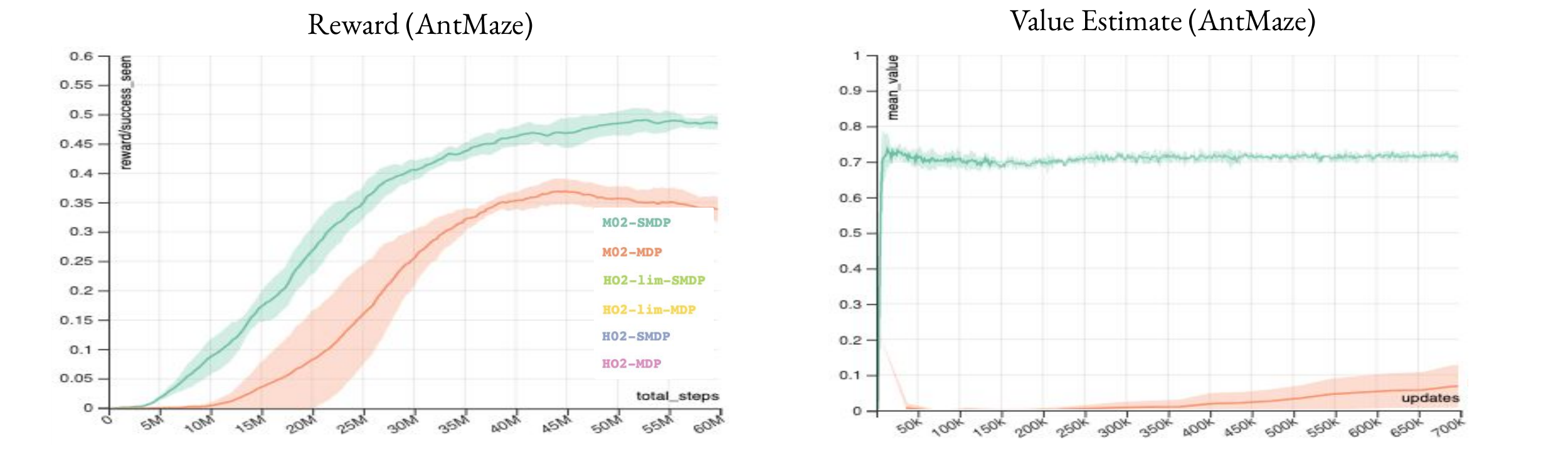}
 \vspace{-3mm}
 \caption{\emph{Semi-MDP vs MDP Learning Curves}. To evaluate the importance of temporal abstraction for policy evaluation and improvement, we compare expected return and value estimates between methods that perform value estimation over the semi-MDP vs MDP, both acting at the option-level. We plot mean (solid line) and standard deviation (shaded region) across 4 seeds. \rebuttal{Curves are periodically obtained by evaluating the current policy on the environment.} As seen by comparing semi-MDP and MDP experiments, contrasting reward and value estimate curves, abstractions reduce value bias, improve value convergence, and policy performance. The degree of improvement directly correlates to the level of temporal compression (MO2 $>$ HO2-lim $>$ HO2). Nevertheless, for the simpler Maze2D domain, temporal abstraction is unnecessary with HO2-MDP performing best. }
 \label{fig:policy_evaluation}
 \vspace{-3mm}
\end{figure}
\vspace{-2mm}
To evaluate the importance of temporal abstractions for learning, specifically \rebuttal{value estimation}, we plot semi-MDP vs MDP policy evaluation learning curves, where \rebuttal{value estimation} occurs either over the semi-MDP or MDP, both acting at the option-level\footnote{The Reward and Value Estimates horizontal axes are aligned despite different metrics due to the asynchronous learner.} (see \Cref{fig:policy_evaluation}). We additionally compare policy performance for both setups, to inspect how important an accurate critic is for policy improvement (see \Cref{fig:policy_evaluation}). Comparing policy performance and evaluation for each method, it is apparent that \rebuttal{value estimation} over MO2's more temporally compressed options yields a faster learning, less biased critic (seen when comparing return and \rebuttal{value estimate} curves). For both domains, improved policy evaluation yields improved policy performance, although the improvement is more significant for the longer horizon AntMaze. Additionally, the more temporally compressed, the larger the improvement for value bias and convergence as well as policy improvement, when \rebuttal{value estimation} occurs over the semi-MDP. This highlights the importance of temporal compression for policy evaluation and improvement. Interestingly, for Maze2D, an increasingly biased critic, HO2-MDP, does not hinder performance, suggesting that here, reasonable advantage estimation is occurring.
\vspace{-2mm}
\subsection{Option-Transition Predictability}
\vspace{-1mm}

In \Cref{sec:offline_bottlenecks}, we proposed discovering \emph{bottleneck options} by optimising $O_{pred} = \sum_{n}\mathbb{E}_{\pi}[\log(p^T(s_{n+1}|s_n, o_n, a_n))]$ (\Cref{eq:o_pred_offline}), predictable option-transitions across offline episodes. As outlined in \Cref{sec:transition_model_learning}, we use a learnt option-transition model $\tilde{p}^T$ to optimise $O_{pred}$, as we do not have access to the true $p^T$. We train such a model using \Cref{eq:O2T_loss}. To validate that \Cref{eq:MO2_loss,eq:O2T_loss} lead to predictable semi-MDP transitions on the \emph{source} domains, despite using a learnt transition model to optimise behaviours, we evaluate the following:
\begin{wraptable}{r}{6cm}
 \vspace{-4mm}
 \scriptsize
 \caption{$CE(\tilde{p}^T)$ (see text)}
 \vspace{1mm}
 \label{tab:cumulative_predictive_error}
 \centering
 \begin{tabular}{lll}
 \toprule
 \multicolumn{1}{c}{Environment} & \multicolumn{1}{c}{Maze2D} & AntMaze \\
 \toprule
 HO2 (offline) & $\mathbf{0.45\pm0.00}$ & $7.20\pm0.01$\\
 HO2-lim (offline) & $1.00\pm0.02$ & $13.00\pm1.00$\\
 MO2 & $0.70\pm0.05$ & $\mathbf{4.00\pm0.05}$\\
 \toprule
 Max (min) & $5.00$ $(0.45)$ & $50.00$ $(4.00)$ \\
 \bottomrule
 \end{tabular}
 \vspace{-2mm}
\end{wraptable}
\vspace{-4mm}
\begin{equation}
CE(\tilde{p}^T) = -\sum_{n=0}^{N} \mathbb{E}_{\substack{s_n, a_n, o_n, s_{n+1} \sim \pi }} [ \log(\tilde{p}^T(s_{n+1}|s_n,o_n,a_n)))]
\label{eq:cum_plan_error}
\end{equation}
The expectation is taken over agent $\pi$ rollouts in the environment after skills are frozen (as opposed to over the offline dataset as in \Cref{eq:o_pred_offline}), but before training on the \emph{transfer} domain. As such, we evaluate predictable semi-MDP transitions according to an agent optimised over the \emph{source} domains, as desired. We use the same notation as in \Cref{sec:mo2_model_sec}. $CE(\tilde{p}^T)$ represents how well our model $\tilde{p}^T$ can predict transitions across the semi-MDPs. This metric is influenced by both the individual transition predictability  and total episodic number of transitions, $N$. We report results in \Cref{tab:cumulative_predictive_error}. The lower the score the more predictable the transitions. In \Cref{tab:cumulative_predictive_error}, the max value refers to the initial value achieved before offline training of $\pi$ and $\tilde{p}^T$ and is shown to contextualise the results. Min refers to minimum value achieved across all methods. Interestingly, for AntMaze, with high dimensional state-action spaces and a long horizon, MO2 leads to the best predictability scores, thanks to its low switch rate transitioning at predictable bottleneck states (\Cref{fig:policy_rollouts}). Conversely, for Maze2D, MO2's abstractions are not necessary for high predictability. This may partly explain why we do not observe \emph{transfer learning} benefits here (\Cref{fig:mo2_return}). For both domains, even though HO2-lim (offline) also reduces option switch rate when compared with HO2 (offline), this reduction does not lead to a lower $CE$, suggesting that penalising option-switching alone does not lead to predictable transitions. Altogether, these results suggest that \Cref{eq:MO2_loss,eq:O2T_loss} lead to compressed temporal abstractions with predictable transitions, with benefits favouring longer horizon, higher dimensional domains (arguably where abstractions are most necessary).

%% file: sections/related_works.tex
\vspace{-2mm}
\section{Related Work}
\vspace{-1mm}
\citet{sutton1999between} introduced the options framework for discovering temporal abstractions. Multiple works built on this framework to improve option switching degeneracy \citep{kamat2020diversity, harb2018waiting}, sample efficiency \citep{wulfmeier2020data, riemer2018learning}, off-policy corrections \citep{levy2017learning, nachum2018data} and optimisation \citep{wulfmeier2020data, Smith2019, Zhang2019DAC}. \citet{wulfmeier2020data} present HO2, a hindsight off-policy options framework combining many of the previous advancements, enabling sample-efficient intra-option learning from off-policy data. \rebuttal{Concurrent to our work, \citet{klissarov2021flexible} published a similar method to HO2 for training all options off-policy.} \revision{While aforementioned methods outperform non-hierarchical equivalents, }\rebuttal{optimising a purely control objective over \emph{source} domains without encouraging options suited for planning, often leads to high-frequency skills that maximise return on \emph{source} domains at the expense of effective exploration and credit-assignment during \emph{transfer}.}

\rebuttal{MO2 is a bottleneck option method similar to  \citet{mcgovern2001automatic,csimcsek2004using, solway2014optimal, harutyunyan2019termination, ramesh2019successor} discovering temporal abstractions that induce plans of minimum description length over a set of tasks \citep{solway2014optimal}. Nevertheless, the following existing approaches do not support continuous state-spaces: \citet{harutyunyan2019termination}, as the predictability objective only supports tabular TD-learning; \citet{solway2014optimal}, as they require constructing a graph over state-space transitions, thus not easily scalable; \citet{csimcsek2004using}, building a partial transition graph with time complexity of $O(T^3)$ ($T =$ horizon length); \citet{mcgovern2001automatic}, discovering regions of maximum diverse density using exhaustive search, thus scaling poorly. Whilst \citet{ramesh2019successor} supports continuous spaces, they are on-policy, discovering successor options using PPO, thus unable to learn from diverse, past experience, commonly available \citep{khetarpal2020towards,caesar2020nuscenes}. In contrast, MO2 is the first bottleneck options approach applied to the offline skill learning setting with continuous state-action spaces.}

Hierarchical and variational approaches present alternate methods for discovering skills. \citet{hausman2018learning, haarnoja2018latent, Wulfmeier2019} discover (multi-level) action abstractions suited for re-purposing, often by encouraging skill distinguishability, but do not support temporal abstractions. \citet{singh2020parrot, merel2018neural, merel2020catch, salter2022priors, goyal2019infobot} encourage temporally consistent behaviours by regularising against auto-regressive, learnt or uninformative priors, yet do not leverage temporal abstractions for \rebuttal{value estimation}. \citet{pertsch2020accelerating, ajay2020opal, co2018self} discover temporally abstract skills, essential for exploration, but predefine their temporal resolution. Unlike bottleneck approaches, prior works do not explicitly optimise for skills beneficial for planning and acting across tasks, potentially limiting their transfer learning benefits.

Similar to MO2 are methods that impose information-theoretic constraints on the discovered skills. \citet{harutyunyan2019termination} introduce the termination critic, encouraging compressed options with a low entropic termination state distribution, $\min\{H(s_f|o)\}$. Unlike MO2, the termination critic \rebuttal{learns options in a tabular setting, and} does not support continuous state-action spaces, limiting their applicability to robotic domains. Furthermore, by building on HO2, MO2 enables intra-option learning, shown to be beneficial. \citet{wang2019learning} present TAIC, that maximises mutual information between options and initial and termination states, $\max\{I(o;s_i, s_f)\}$, while simultaneously minimising individual mutual informations, $\min\{I(o;s_i) + I(o;s_f)\}$, thereby encouraging options independent of their initial and termination states. \citet{wang2019learning} learn their abstractions via a variational auto-encoder approach. Unlike MO2, TAIC is on-policy and builds on PPO \citep{schulman2017proximal} which reduces sample efficiency.  Additionally, unlike MO2, TAIC's options have limited ability to specialise, due to their independence to initial and termination states. \citet{shiarlis2018taco} introduce TACO, a weakly supervised approach for discovering temporally compressed abstractions aligning with sub-tasks, given a sub-task sketch by the user. MO2 discovers skills unsupervised presenting a more viable approach in many scenarios where supervision is expensive.  

Finally, in contrast to MO2, \rebuttal{\citet{machado2017laplacian} propose eigen-, instead of bottleneck-, options, with superior performance in \emph{online} settings when the goal is far away from the \emph{bottleneck states} (e.g. room corners for the 4-room domain). Eigen-options minimise diffusion time of an agent across the environment. We consider extending the eigen-options framework to the \emph{offline} setting as future work.} Similarly, approaches like \citet{jinnai2019exploration} discover options (or latent actions) with wide state coverage \citep{eysenbach2018diversity, gregor2016variational, achiam2018variational, kamat2020diversity}. These works take the intrinsic-curiosity RL paradigm. Whilst encouraging diversity is important for skill development when learning \emph{online} from scratch (over sparse rewards), it is unclear whether these skills are temporally compressed and suited for \rebuttal{value estimation}. We consider extending MO2 to the online skill learning context, \rebuttal{and learning from sub-optimal \emph{source domain} demonstrations \citep{peng2019advantage},} as future work.

%% file: sections/conclusion.tex
\vspace{-4mm}
\section{Conclusions}
\vspace{-3mm}
We introduce Model-Based Offline Options, MO2, a novel offline hindsight bottleneck options framework supporting continuous state-action spaces, that discovers skills suited for planning (in the form of \rebuttal{value estimation}) and acting across modular tasks (MDPs whose optimal policies are obtained by recomposing shared temporal abstractions). MO2 achieves this with a \emph{predictability} objective encouraging predictable option-terminations across an episode. Once options are learnt offline, we freeze them and perform online transfer learning over the option-induced semi-MDP, learning and acting over the temporally abstract skill space. We compare MO2 against state-of-the-art baselines on complex continuous control domains and perform an extensive ablation study. We demonstrate that MO2's options are bottleneck aligned and improve acting (and exploring), \rebuttal{value estimation}, and learning of a jumpy option-transition model. On the challenging, sparse, long-horizon, AntMaze domain, MO2 drastically outperforms all baselines.

%% file: sections/appendix.tex
\newpage
\appendix
\section*{{\LARGE Appendix}}

We provide the reader with experimental details required to reproduce the results in the main paper.

\section{Architecture}

We continue by outlining the model architectures across domains and experiments. Each policy network ($\pi^C, \pi^L, \beta$) and the option-transition model $\tilde{p}^T$ are comprised of a feedforward module outlined in \Cref{primary-building-block-setup}. $\pi^C$ then has a component layer of size $128$ for each option.
We apply a tanh or softmax activations over network outputs, where appropriate, to match the predefined output ranges of any given module. The critic, $Q$, is also a feedforward module. For $\pi^C$ we use a categorical latent space of size $4$ (number of options). We find this dimensionality suffices for expressing the diverse behaviours in our domains. $\tilde{p}^T$ is also a gaussian mixture model with 4 heads. On the transfer experiments we freeze $\pi^L, \beta$ and train a newly instantiated $\pi^C$ to recompose prior skills. $\tilde{p}^T$ is not used during online transfer.

\begin{table}
  \caption{Feedforward Module, $\pi^{\{C, L\}}, \beta, \tilde{p}^T, Q$}
  \label{primary-building-block-setup}
  \centering
  \begin{tabular}{ll}
    \toprule
    hidden layers & $(256, 256)$ \\
    hidden layer activation & elu \\
    output activation & linear\\
    \bottomrule
  \end{tabular}
\end{table}

\section{Algorithmic Details}
\label{app:alg_details}

\begin{table}
\caption{Offline Skill Discovery Setup}
\centering
\label{tab:bc-setup}
  \begin{tabular}{lll}
    \toprule
    \multicolumn{1}{c}{Environment} & \multicolumn{1}{c}{Maze2D} & AntMaze                 \\
    \midrule
    $\pi^{\{C, L\}}, \beta$ learning rate & $3e^{-4}$ & $3e^{-4}$ \\
    $\tilde{p}^T$ learning rate & $3e^{-4}$ & $3e^{-4}$ \\
    number of options & $4$ & $4$ \\
    $\alpha$ predictability weight & $1.0$ & $0.2$ \\
    $m$ $\log(\tilde{p}^T)$ offset & $13$ & $103$ \\
    batch size & $256$ & $128$ \\
    sequence length & $100$ & $100$ \\
    \bottomrule
  \end{tabular}
\end{table}

\begin{table}
\fontsize{8}{10}\selectfont
\caption{Online Transfer Learning Setup}
\centering
\label{tab:rl-setup}
  \begin{tabular}{llllllllllllll}
    \toprule
    Environment & Maze2D & AntMaze  \\
    \midrule
    $Q$ learning rate & $1e^{-4}$ & $1e^{-4}$ \\
    $\pi^C$ learning rate & $1e^{-4}$ & $1e^{-4}$ \\
    number options & $4$ & $4$ \\
    $\epsilon$ & $1$ & $1$ \\
    $\epsilon_{KL}$ & $1$ & $1$ \\
    data-grad ratio & $5000$ & $50$ \\
    batch size & $256$ & $256$ \\
    sequence length & $10$ & $10$\\
    \bottomrule
  \end{tabular}
\end{table}

\subsection{Offline Skill Discovery}
\label{app:offline_skill_discovery}
In \Cref{eq:MO2_loss} we introduce MO2's objective $O_{mo2}$ for discovering bottleneck options. In practice, we actually use the following slightly modified objective:

\begin{equation}
\underset{\pi^L, \pi^C, \beta}{\max}O_{mo2}(\pi^L, \pi^C, \beta) = \mathbb{E}_{\substack{s_t, a_t, h_t \sim D \\ o \sim \pi^H(\cdot|h_t), a \sim \pi^L(\cdot|s_t, o) \\ s_f \sim \tilde{p}^T(\cdot|s_t, o, a)}} [\alpha\beta(s_t|h_t)\log(\tilde{p}^T(s_f|s_t,o,a) - m) + \log(\pi(a_t, o|h_t))]
\label{eq:MO2_loss_detailed}
\end{equation}

$m$ and $\alpha$ are hyperparameters. $m$ is set to the maximum possible $\tilde{p}^T(s_f|s_t,o,a)$ value and is a function of the minimum permitted covariance $\Sigma$ of our model $\tilde{p}^T$ ($\sigma$ set to $1e^{-3}$ for $\Sigma = \sigma \mathbb{I}$). As such, $m$ ensures $O_{pred}$ is always negative, thus encouraging minimal switches that align with bottleneck states, as outlined in \Cref{fig:mo2_model}. $\alpha$ is set using grid search (specifically [$ 1e^{-1}, 2e^{-1}, 4e^{-1}, 6e^{-1}, 8e^{-1}, 1e^{0}$]), and weighs the importance of the two objects $O_{pred}$ and $O_{bc}$. \rebuttal{The offline data collected by behavioural policy, $b(a|h)$, consists of tuples of experience in the form $(s_t, a_t)$.}

We train the options framework ($\pi^C, \pi^L, \beta$) and option-transition model $\tilde{p}^T$ in parallel using \Cref{eq:MO2_loss_detailed} and \Cref{eq:O2T_loss} respectively. Algorithmic hyperparameter details are outlined in \Cref{tab:bc-setup}. We run four seeds. During transfer, we select the seed that traverses the maze deepest. We note, however, that there is minimal difference between each seed. \emph{Batch size} refers to the number of independently sampled trajectory snippets from the offline dataset, each of length \emph{sequence length}. We sample these batches uniformly, according to \citet{wulfmeier2020data}. We run experiments until $O_{mo2}$ convergence ($1e^{6}$ gradients). For the baselines, \emph{HO2 (offline), HO2-lim (offline)}, we set $\alpha$ to zero, yet still train $\tilde{p}^T$ for evaluation purposes. The other hyperparameters are kept constant. Finally, for \emph{HO2-lim (offline)}, we run a sweep over $N$ (specifically [$1, 5, 10$]), the number of permitted option switches, when training on Equation (17) from the Appendix of \citet{wulfmeier2020data}. We report results for $N = 1$, as this is the experiment that achieves the lowest switch rate.

\subsection{Online Transfer Learning}
\label{app:online_transfer}

During this stage of training we freeze the option-policies ($\pi^L, \beta$). The categorical option controller, $\pi^C$ is initialized randomly on the transfer task (inline with \citet{wulfmeier2020data}) and is trained to reorder options to solve the task. \rebuttal{We note that while some skill transfer approaches \citep{salter2022priors, pertsch2020accelerating} note performance gains by additionally transferring $\pi^C$, we did not observe this.} The RL setup is akin to the CMPO \citep{neunert2020continuous}. We note any changes in \Cref{tab:rl-setup}. Hyper-parameter sweeps are performed for \emph{data collection to gradients ratio} (data-grad ratio) (specifically [$5e^{1}, 5e^{2}, 5e^{3}$]) for each method as we find this makes a significant difference. We report the most performant setting for each approach. For HO2-scratch, we train an HO2 agent from scratch, using the same architecture as the other methods, and use the default setting/hyperparameters in \citet{wulfmeier2020data}.

\subsubsection{MDP learning}

For this setup, the agent acts in the environment at the option-level. Learning occurs at the original MDP level, however. The following, per timestep, tuple of experiences are stored in the replay-buffer: $\{s_t, o_t, r_t, s_{t+1}\}$. We use these samples to train the critic and policy, using CMPO. We perform temporal-difference TD(0) learning \citep{sutton1988learning} for training the critic. Crucially, this occurs per timestep, at the original temporal resolution of the MDP. As such, the option's temporal abstractions are not used to perform more efficient TD learning across long horizons.

\subsubsection{Semi-MDP learning}

For this setup, the agent acts in the environment at the option-level. Learning also occurs at the option-level (over the option-induced semi-MDP). The following tuple of experiences are stored in the replay-buffer: $\{s_t, o_t, g_t, s_{t+k}\}$, where $k$ is specified by the option's sampled termination condition in the environment. $s_{t}$ represents the option's $o_t$ initiation state, $s_{t+k}$ the termination state, and $g_t = \sum_{i = t}^{t + k}\gamma^{i - t}r_i$ the discounted cumulative rewards during its execution. We use these samples to train the critic and policy, using CMPO. We perform temporal-difference TD(0) learning \citep{sutton1988learning} for training the critic. Crucially, this occurs at the option-level, over the semi-MDP. As such, the option's temporal abstractions are used to perform more efficient TD learning across long horizons.

\section{Environments}
\label{app:environments}
In this section, \revision{we describe in detail the D4RL \citep{fu2020d4rl} domains used in our experiments.} 

\subsection{Maze2D}

\revision{

For this D4RL domain \citep{fu2020d4rl} a 2D pointmass agent must reach a fixed goal location. We test our algorithm on the `maze2D-large' variant with the maze layout as depicted in \Cref{fig:d4rl_setup}. 

\subsubsection{Source Domains}

\begin{wrapfigure}{r}{0.67\textwidth}
  \vspace{-5mm}
  \begin{center}
    \includegraphics[trim={6mm 8mm 6mm 2mm},clip,width=0.62\textwidth]{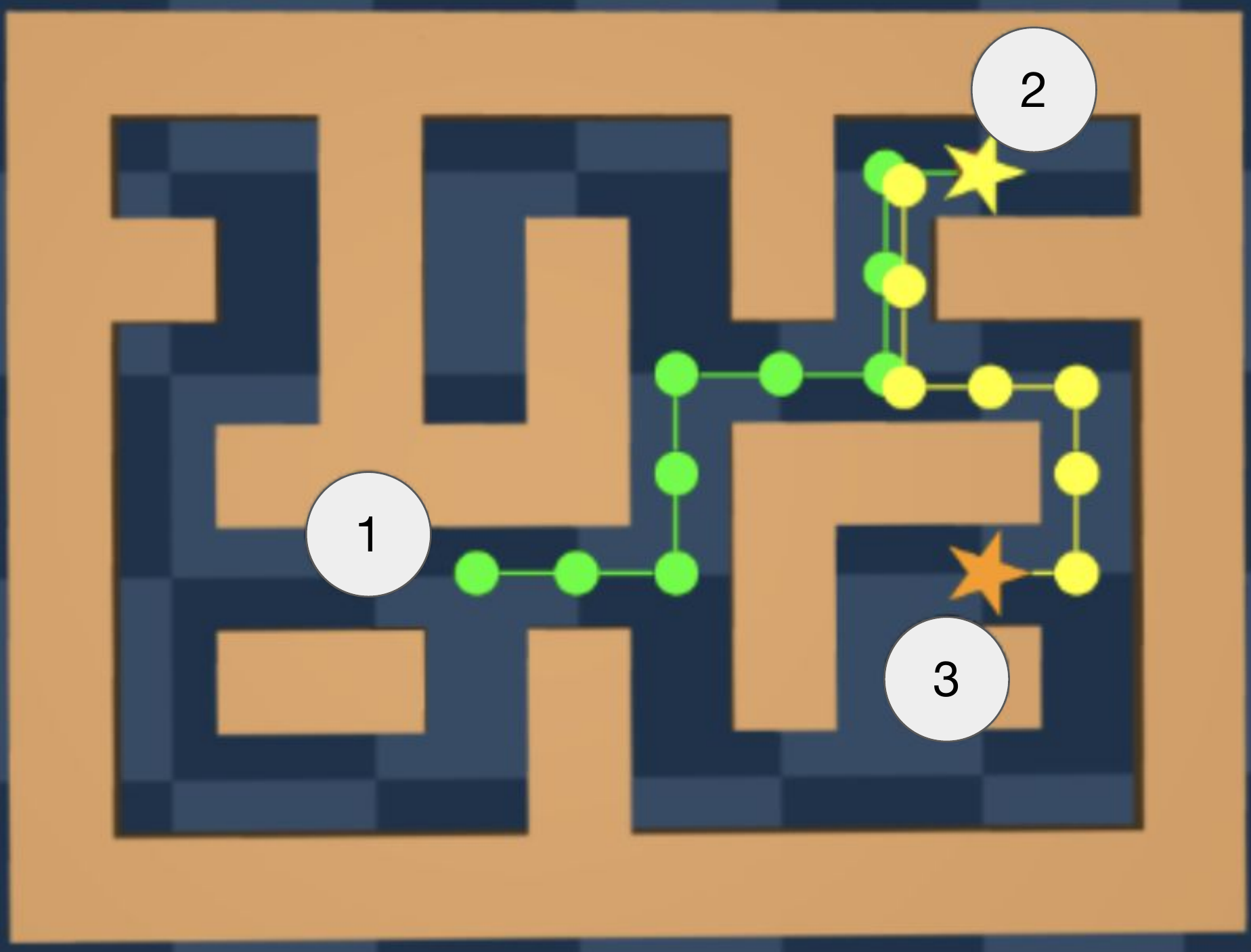}
  \end{center}
  \vspace{-3mm}
  \caption{\revision{\emph{Offline Data Generation Example.} The agent randomly spawns in (1). Goal location (2) is then randomly sampled (denoted by a star). A planner generates waypoints to reach the goal (shown as circles). For \emph{Maze2D}, a PD controller is used to follow the waypoints. For \emph{AntMaze}, a trained goal-reaching policy \citep{fu2020d4rl} is used to follow them. Once the goal location (2) is reached, the next goal location (3) is randomly sampled and (2) becomes the spawning location. The maze layout remains static, as displayed, between \emph{source} and \emph{transfer} domains. During \emph{transfer}, an unknown goal location must be reached. See text for further details. Figure modified from \citet{fu2020d4rl}.}}
  \label{fig:d4rl_setup}
\end{wrapfigure}

The offline data $D$ is generated by randomly selecting goal locations in the maze and then using a planner that generates sequences of waypoints (as shown in \Cref{fig:d4rl_setup}) that are followed using a PD controller to reach the goal. Once the goal is reached, a new goal is sampled and the process continues. Distinct goals require distinct corridor traversals. As such, corridor intersections represent the bottleneck states for these source domains. The maze layout remains static and is not randomised between tasks. The offline dataset contains $4 * 10^6$ environment transitions (in the form of $(s,a)$ tuples). The state-space corresponds to 2D Cartesian coordinates of the agent and the actions correspond to its velocity, clipped within the range of $(-1, 1)$ per dimension. Crucially, goal location is not provided to the agent, as the agent must learn task agnostic behaviours. This dataset is publically available from \citet{fu2020d4rl}.

\subsubsection{Transfer Domains}

We evaluate our algorithms on a modified version of `maze2d-eval-large' from \citet{fu2020d4rl}. The maze layout remains the same as the source domains'. The goal location now remains static, is not randomised, and the agent must discover its location and how to consistently reach it as quickly as possible.} The original Maze2D transfer domain spawns the agent (at the start of the episode) uniformly across the maze. The agent is rewarded only upon goal reaching ($+1$). The agent sometimes spawns close to the goal, other times far away. As such, credit assignment and exploration are not particularly problematic as the spawn locations provide a form of curriculum for learning values and exploring the environment across all starting locations in the maze. Therefore, this domain does not necessitate temporal abstractions for acting and learning. To increase the difficulty, we modify the environment such that the agent always spawns at the furthest distance from the goal (with respect to steps required to reach target). Now, abstractions are more important for credit assignment and exploration. Nevertheless, the state-action space for this domain is low dimensional, and the episodic horizon is not long, meaning the task is still not particularly sparse. Additionally, to keep with the modular task setting (as detailed in \Cref{sec:intro}), we terminate the episode early when the goal is reached (unlike in the original domain). Below we detail the maze layout for our modified setting:

\begin{lstlisting}
// Modified Maze2D
// R = spawn location, G = goal location, 1 = walls, 0 = empty space

Maze2D = [[1, 1, 1, 1, 1, 1, 1, 1, 1, 1, 1, 1],
            [1, 0, 0, 0, 0, 1, 0, 0, 0, 0, 0, 1],
            [1, 0, 1, 1, 0, 1, 0, 1, 0, 1, 0, 1],
            [1, 0, 0, 0, 0, 0, 0, 1, 0, 0, 0, 1],
            [1, 0, 1, 1, 1, 1, 0, 1, 1, 1, 0, 1],
            [1, 0, 0, 1, 0, 1, 0, 0, 0, 0, 0, 1],
            [1, 1, 0, 1, 0, 1, 0, 1, 0, 1, 1, 1],
            [1, R, 0, 1, 0, 0, 0, 1, 0, 0, G, 1],
            [1, 1, 1, 1, 1, 1, 1, 1, 1, 1, 1, 1]]

\end{lstlisting}

\subsection{AntMaze}

\revision{

This D4RL domain \citep{fu2020d4rl} is a navigation domain that replaces the 2D pointmass from Maze2D with a complex 8-DoF `Ant' quadraped robot. This domain is introduced to mimic more challenging real-world robotic nagivation tasks. We test our algorithm on `ant-large-diverse' variant using the same maze layout as depicted in \Cref{fig:d4rl_setup}.

\subsubsection{Source Domains}

The offline data $D$ is created by training a goal reaching policy and using it in conjunction with the same high-level waypoint generator from Maze2D to provide sub-goals that guide the agent to the goal. For the `ant-large-diverse' dataset, the ant is commanded to a random goal location from a random start location. Both locations are randomly sampled across the maze. Akin to Maze2D, distinct goals require distinct corridor traversals. As such, corridor intersections represent the bottleneck states for these source domains. The maze layout remains static and is not randomised between tasks. The offline dataset contains $10^6$ environment transitions (in the form of $(s,a)$ tuples). The state-space is $111$-dimensional corresponding to position and velocity of each joint, and contact forces. The action space is a $8$-dimensional continuous space corresponding to the torque applied to each actuator, clipped between $(-1, 1)$ per dimension. Crucially, goal location is not provided to the agent, as the agent must learn task agnostic behaviours. This dataset is publically available from \citet{fu2020d4rl}.

\subsubsection{Transfer Domains}

We evaluate our algorithms on a modified version of `ant-eval-large-diverse' from \citet{fu2020d4rl}. The maze layout remains the same as the source domains'. The goal location now remains static, is not randomised, and the agent must discover its location and how to consistently reach it as quickly as possible. Unlike the source domains, the agent's spawning location is not randomised between episodes and consistently spawns in the corner of the maze. The agent is rewarded only upon goal reaching ($+1$).} For this transfer domain, we experimentally find that reaching the original target location is tricky for all methods. Whilst MO2 strongly outperforms all other approaches, its success rate is still low (within the allocated online training budget). We suspect this may be an issue with the number of (sub-)trajectories offline reaching the goal location. All our methods discover skills by maximising the log-likelihood of offline data. If the goal is rarely reached offline, neither method will favour discovering skills that reach it. As such, we change the goal location to a slightly closer location, more representative of offline behaviours (which we visually inspected). We used the following maze layout:

\begin{lstlisting}
// Modified AntMaze
// R = spawn location, G = goal location, 1 = walls, 0 = empty space

AntMaze = [[1, 1, 1, 1, 1, 1, 1, 1, 1, 1, 1, 1],
            [1, R, 0, 0, 0, 1, 0, 0, 0, 0, 0, 1],
            [1, 0, 1, 1, 0, 1, 0, 1, 0, 1, 0, 1],
            [1, 0, 0, 0, 0, 0, 0, 1, 0, 0, 0, 1],
            [1, 0, 1, 1, 1, 1, 0, 1, 1, 1, 0, 1],
            [1, 0, 0, 1, 0, 1, 0, 0, 0, 0, 0, 1],
            [1, 1, 0, 1, 0, 1, 0, 1, 0, 1, 1, 1],
            [1, 0, 0, 1, 0, 0, G, 1, 0, 0, 0, 1],
            [1, 1, 1, 1, 1, 1, 1, 1, 1, 1, 1, 1]]

\end{lstlisting}

\rebuttal{\section{Proofs and Discussions}
\label{sec:all_proofs}

We continue by highlighting why MO2's objective leads to bottleneck state discovery. We provide proofs coupled with explanations.

\subsection{Why does \Cref{eq:o_pred} lead to bottleneck options?}
\label{sec:why_bottlenecks}

We continue by demonstrating why \Cref{eq:o_pred} leads to the minimal number of low entropic, transition state distributions, able to reconstruct offline behaviours. To achieve this we first demonstrate why maximising \Cref{eq:o_pred} is equivalent to minimising the conditional entropy of plans. $\mathbb{H}_{\pi}[s_{0:N}|o_{0:N}, a_{0:N}]$ represents the conditional entropy of plans over sequential option-transition states $s_{0:N}$ conditioned on sequential options $o_{0:N}$ and their corresponding first initiated action $a_{0:N}$.
\begin{equation}
\begin{split}
 \mathbb{H}_{\pi}[s_{0:N}|o_{0:N}, a_{0:N}] &= - \mathbb{E}_{s_{0:N}, o_{0:N}, a_{0:N} \sim \pi}[\log(p_{\pi}(s_{0:N}|o_{0:N}, a_{0:N}))]  \\ 
 &= - \mathbb{E}_{s_{0:N}, o_{0:N}, a_{0:N} \sim \pi}[\prod_{n=0}^{N}\log(p^T(s_{n+1}|s_n, o_n, a_n))]  \\ 
 &= - \sum_{n=0}^N\mathbb{E}_{s_{0:N}, o_{0:N}, a_{0:N} \sim \pi}[\log(p^T(s_{n+1}|s_n, o_n, a_n))]  \\ 
 &= - \sum_{n=0}^N\mathbb{E}_{s_n, o_n, a_n, s_{n+1} \sim \pi}[\log(p^T(s_{n+1}|s_n, o_n, a_n))]  \\  
&= \sum_{n=0}^N \mathbb{H}_{\pi}[s_{n+1}|s_{n}, o_{n}, a_{n}] \\
&:= - O_{pred}  \; \text{(\Cref{eq:o_pred})}
\label{eq:o_pred_entropy_proof}   
\end{split}
\end{equation}
We subscript $\mathbb{H}$ with $\pi$ to emphasise that this entropy (and plans) is dependent on our option-policy $\pi$ that defines the option-transition model $p^T(s_{n+1}|s_n, o_n, a_n)$ and how options and actions are chosen (from $\pi^C(o|s)$ and $\pi^L(a| s, o)$). $s_{0:N}, o_{0:N}, a_{0:N} \sim \pi$ represents that the expectation over sequential transition states, options and actions are sampled from our policy $\pi$ acting in the environment (i.e. from $p_{\pi}(s_{0:N}, o_{0:N}, a_{0:N})$).

The second line of \Cref{eq:o_pred_entropy_proof} holds due to the product rule and independence between $s_{n+1}$ and all other variables after conditioning on $s_n, o_n, a_n$, given our graphical model of the agent $\pi$ and environment. The fourth line holds due to the integral of all extraneous variables, that $p^T$ is not dependent on, integrating to 1. The final line is equivalent to the fourth by the definition of \Cref{eq:o_pred}.

We therefore demonstrate that maximising the $O_{pred}$ objective in \Cref{eq:o_pred} is equivalent to minimising the conditional entropy of plans over sequential option-transition states $\mathbb{H}_{\pi}[s_{0:N}|o_{0:N}, a_{0:N}]$ as well as the cumulative sum of individual conditional option-transition entropies $\sum_{n=0}^N \mathbb{H}_{\pi}[s_{n+1}|s_{n}, o_{n}, a_{n}]$. If individual entropies are enforced to be positive (achievable as shown in \Cref{app:alg_details}) then to minimise \Cref{eq:o_pred_entropy_proof}, transitions should be as sparse as possible as to minimise the accumulation of positive option-transition entropies. 

If we assume for now that $\pi(a|h) = b(a|h)$ (with $b$ representing the behavioural distribution that created the offline data), and constant environment dynamics, then the distribution over trajectories induced by $\pi$ equates to the offline data distribution over the \emph{source} domains. As such, \Cref{eq:o_pred_entropy_proof} encourages a decomposition of offline behaviours into compressed temporal abstractions between sequential decision states able to reconstruct offline behaviours by their recomposition (though options) with high confidence (low $\mathbb{H}_{\pi}[s_{0:N}|o_{0:N}, a_{0:N}]$). Additionally, individual option-transitions entropies $\mathbb{H}_{\pi}[s_{n+1}|s_{n}, o_{n}, a_{n}]$ should be low, thus aligning with bottleneck states as depicted in \Cref{fig:mo2_model}. Altogether, \Cref{eq:o_pred} leads to the \emph{optimal-compression (bottleneck state alignment)} scenario in \Cref{fig:mo2_model}.

In practice, $\pi(a|h) \neq b(a|h)$. However, the $O_{bc}$ objective in \Cref{eq:MO2_loss,eq:MO2_loss_detailed} encourages both to align with each other. The $\alpha$ hyper-parameter in \Cref{eq:MO2_loss_detailed} is set low enough such that the $O_{pred}$ component of MO2's objective does not hinder $\pi$'s ability to align with $b$. Thus, over time, as $\pi$ distils the offline behaviours, \Cref{eq:MO2_loss,eq:MO2_loss_detailed} will discover bottleneck options able to reconstruct the offline modular behaviours across the \emph{source} MDPs.

\subsection{Under what assumptions does optimising $O_{pred}$ from \Cref{eq:o_pred} $=$ \Cref{eq:o_pred_offline}?}
\label{sec:why_both_o_pred_terms_equivalent}

\Cref{eq:MO2_loss,eq:MO2_loss_detailed} do not use the form of $O_{pred}$ from \Cref{eq:o_pred} but instead the alternate definition in \Cref{eq:o_pred_offline}. We continue by highlighting under what assumptions optimising either form is equivalent.

Our final proof relies on two concepts. Firstly, instead of considering individual $n^{th}$ option initiation state distributions $p_{\pi}(s_n)$ across the episode, we consider the stationary option initiation state distribution $p_{\pi}(s_i)$ by marginalising over the transition number dimension $n$:
\begin{equation}
  p_{\pi}(s_i) = E_{n \sim \pi}[p_{\pi}(s_n)]
  \label{eq:marg_n}
\end{equation}
$s_i$ denotes the marginal initiation state distribution, independent of option transition number $n$. The distribution over $n$ is dependent on $\pi$ which is why $n \sim \pi$ in the expectation. The second concept that we rely on is how to obtain the same marginal initiation state distribution by marginalising over the $t$ rather than $n$. As highlighted in \Cref{sec:offline_bottlenecks} we have a closed form solution to a related quantity $\beta(s_t|h_t)$ that represents the probability that $s_t$ is initial given history $h_t$. Given this quantity, we can calculate $p_{\pi}(s_i)$ by marginalising across $t$ and $h_t$ as follows:
\begin{equation}
  p_{\pi}(s_i) = E_{t \sim U\{0, T\}, h_t \sim \pi} [\beta(s_t|h_t)p_{\pi}(s_t|h_t)]
  \label{eq:marginal_time}
\end{equation}
$U\{0, T\}$ denotes a uniform categorical distribution between 0 and $T$, the episodic length. $p_{\pi}(s_t|h_t)$ denotes the probability state distribution at timestep $t$ given history $h_t$, for policy $\pi$.
\Cref{eq:marginal_time} without the $\beta(s_t|h_t)$ term corresponds to the stationary state visitation distribution of policy $\pi$ (e.g. $p_{\pi}(s) = E_{t \sim U\{0, T\}, h_t \sim \pi}[p_{\pi}(s_t|h_t)]$ which is independent of $t$). The additional weights weigh how probable that any $s_t$ is an option initiation state given $h_t$ according to our option-policy $\pi$. We continue by using \Cref{eq:marg_n,eq:marginal_time} to show to relation between \Cref{eq:o_pred,eq:o_pred_offline}:
\begin{equation}
\begin{split}
O_{pred}  \; \text{(\Cref{eq:o_pred})} &= \sum_{n=0}^N\mathbb{E}_{s_n, o_n, a_n, s_{n+1} \sim \pi}[\log(p^T(s_{n+1}|s_n, o_n, a_n))] \\ &= 
N \mathbb{E}_{\substack{s_i \sim \pi \\ o \sim \pi^C(\cdot|s_i), a \sim \pi^L(\cdot | s_i, o) \\ s_f \sim p^T(\cdot|s_i, o, a)}}[\log(p^T(s_{f}|s_i, o, a))] \\ &\propto
\mathbb{E}_{\substack{s_i \sim \pi \\ o \sim \pi^C(\cdot|s_i), a \sim \pi^L(\cdot | s_i, o) \\ s_f \sim p^T(\cdot|s_i, o, a)}}[\log(p^T(s_{f}|s_i, o, a))] \\ &= \mathbb{E}_{\substack{t \sim U\{0, T\}, s_t, h_t \sim \pi \\ o \sim \pi^H(\cdot|h_t), a \sim \pi^L(\cdot|s_t, o) \\ s_f \sim p^T(\cdot|s_t, o, a)}}[\beta(s_t|h_t)\log(p^T(s_{f}|s_t, o, a))] \\
&= \mathbb{E}_{\substack{s_t, h_t \sim D \\ o \sim \pi^H(\cdot|h_t), a \sim \pi^L(\cdot|s_t, o) \\ s_f \sim p^T(\cdot|s_t, o, a)}}[\beta(s_t|h_t)\log(p^T(s_{f}|s_t, o, a))] \; \text{if $\pi(a|h) = b(a|h)$ } \\
&:=  O_{pred} \; \text{(\Cref{eq:o_pred_offline})} 
\label{eq:eq_3_is_eq_4_proof}
\end{split}
\end{equation}
The second line of \Cref{eq:eq_3_is_eq_4_proof} is equivalent to the first by marginalising over the depth dimension $n$, akin to \Cref{eq:marg_n}, and then grouping into marginal initial transition state distribution $s_i \sim \pi$ of options and their conditional terminal state distribution $s_f \sim p^T(\cdot|s_i, o, a)$. $s_i \sim \pi$ is shorthand for sampling from $p_{\pi}(s_i)$. In the expectation, we explicitly report how options and actions are sampled. $N$ represents the expected number of transitions across episodes. The third line is proportional to the second by a factor of $N$. The fourth line is obtained by applying \Cref{eq:marginal_time} and grouping $p_{\pi}(s_t|h_t)$ into the expectation (i.e. $s_t \sim \pi$). 

In practice, we do not have access to $s_t$ and  $h_t$ samples from option-policy $\pi$ and thus cannot calculate the expectation in the fourth line. Nevertheless, if we assume $\pi(a|h) = b(a|h)$, then we can use offline data samples (from $D$) instead. As such, we obtain the fifth line from \Cref{eq:eq_3_is_eq_4_proof}. We omit $t \sim U\{0, T\}$ for simplicity and to keep notation consistent with \Cref{sec:offline_bottlenecks}. However, we are still sampling $t$ as aforementioned. The final line holds by definition. We have therefore proved that \Cref{eq:o_pred} $\propto$ \Cref{eq:o_pred_offline} under the assumption that $\pi(a|h) = b(a|h)$. Proportionality does not influence optimisation and therefore both are equivalent from a learning perspective. As mentioned in \Cref{sec:why_bottlenecks}, $\pi(a|h) \neq b(a|h)$. However, $O_{bc}$ encourages both to align over time, ensuring $O_{pred}$ leads to bottleneck options.}

%% file: collas2022_conference.bbl
\begin{thebibliography}{52}
\providecommand{\natexlab}[1]{#1}
\providecommand{\url}[1]{\texttt{#1}}
\expandafter\ifx\csname urlstyle\endcsname\relax
  \providecommand{\doi}[1]{doi: #1}\else
  \providecommand{\doi}{doi: \begingroup \urlstyle{rm}\Url}\fi

\bibitem[Abdolmaleki et~al.(2018)Abdolmaleki, Springenberg, Tassa, Munos,
  Heess, and Riedmiller]{abdolmaleki2018maximum}
Abbas Abdolmaleki, Jost~Tobias Springenberg, Yuval Tassa, Remi Munos, Nicolas
  Heess, and Martin Riedmiller.
\newblock Maximum a posteriori policy optimisation.
\newblock \emph{arXiv preprint arXiv:1806.06920}, 2018.

\bibitem[Achiam et~al.(2018)Achiam, Edwards, Amodei, and
  Abbeel]{achiam2018variational}
Joshua Achiam, Harrison Edwards, Dario Amodei, and Pieter Abbeel.
\newblock Variational option discovery algorithms.
\newblock \emph{arXiv preprint arXiv:1807.10299}, 2018.

\bibitem[Ajay et~al.(2020)Ajay, Kumar, Agrawal, Levine, and
  Nachum]{ajay2020opal}
Anurag Ajay, Aviral Kumar, Pulkit Agrawal, Sergey Levine, and Ofir Nachum.
\newblock Opal: Offline primitive discovery for accelerating offline
  reinforcement learning.
\newblock \emph{arXiv preprint arXiv:2010.13611}, 2020.

\bibitem[Bacon et~al.(2017)Bacon, Harb, and Precup]{Bacon2017}
Pierre~Luc Bacon, Jean Harb, and Doina Precup.
\newblock {The option-critic architecture}.
\newblock In \emph{31st AAAI Conference on Artificial Intelligence, AAAI 2017},
  pp.\  1726--1734, 2017.

\bibitem[Cabi et~al.(2019)Cabi, Colmenarejo, Novikov, Konyushkova, Reed, Jeong,
  Zolna, Aytar, Budden, Vecerik, et~al.]{cabi2019scaling}
Serkan Cabi, Sergio~G{\'o}mez Colmenarejo, Alexander Novikov, Ksenia
  Konyushkova, Scott Reed, Rae Jeong, Konrad Zolna, Yusuf Aytar, David Budden,
  Mel Vecerik, et~al.
\newblock Scaling data-driven robotics with reward sketching and batch
  reinforcement learning.
\newblock \emph{arXiv preprint arXiv:1909.12200}, 2019.

\bibitem[Caesar et~al.(2020)Caesar, Bankiti, Lang, Vora, Liong, Xu, Krishnan,
  Pan, Baldan, and Beijbom]{caesar2020nuscenes}
Holger Caesar, Varun Bankiti, Alex~H Lang, Sourabh Vora, Venice~Erin Liong,
  Qiang Xu, Anush Krishnan, Yu~Pan, Giancarlo Baldan, and Oscar Beijbom.
\newblock nuscenes: A multimodal dataset for autonomous driving.
\newblock In \emph{Proceedings of the IEEE/CVF conference on computer vision
  and pattern recognition}, pp.\  11621--11631, 2020.

\bibitem[Co-Reyes et~al.(2018)Co-Reyes, Liu, Gupta, Eysenbach, Abbeel, and
  Levine]{co2018self}
John Co-Reyes, YuXuan Liu, Abhishek Gupta, Benjamin Eysenbach, Pieter Abbeel,
  and Sergey Levine.
\newblock Self-consistent trajectory autoencoder: Hierarchical reinforcement
  learning with trajectory embeddings.
\newblock In \emph{International Conference on Machine Learning}, pp.\
  1009--1018. PMLR, 2018.

\bibitem[Dasari et~al.(2019)Dasari, Ebert, Tian, Nair, Bucher, Schmeckpeper,
  Singh, Levine, and Finn]{dasari2019robonet}
Sudeep Dasari, Frederik Ebert, Stephen Tian, Suraj Nair, Bernadette Bucher,
  Karl Schmeckpeper, Siddharth Singh, Sergey Levine, and Chelsea Finn.
\newblock Robonet: Large-scale multi-robot learning.
\newblock \emph{arXiv preprint arXiv:1910.11215}, 2019.

\bibitem[Eysenbach et~al.(2018)Eysenbach, Gupta, Ibarz, and
  Levine]{eysenbach2018diversity}
Benjamin Eysenbach, Abhishek Gupta, Julian Ibarz, and Sergey Levine.
\newblock Diversity is all you need: Learning skills without a reward function.
\newblock \emph{arXiv preprint arXiv:1802.06070}, 2018.

\bibitem[Fu et~al.(2020)Fu, Kumar, Nachum, Tucker, and Levine]{fu2020d4rl}
Justin Fu, Aviral Kumar, Ofir Nachum, George Tucker, and Sergey Levine.
\newblock D4rl: Datasets for deep data-driven reinforcement learning.
\newblock \emph{arXiv preprint arXiv:2004.07219}, 2020.

\bibitem[Goyal et~al.(2019)Goyal, Islam, Strouse, Ahmed, Botvinick, Larochelle,
  Bengio, and Levine]{goyal2019infobot}
Anirudh Goyal, Riashat Islam, Daniel Strouse, Zafarali Ahmed, Matthew
  Botvinick, Hugo Larochelle, Yoshua Bengio, and Sergey Levine.
\newblock Infobot: Transfer and exploration via the information bottleneck.
\newblock \emph{arXiv preprint arXiv:1901.10902}, 2019.

\bibitem[Gregor et~al.(2016)Gregor, Rezende, and
  Wierstra]{gregor2016variational}
Karol Gregor, Danilo~Jimenez Rezende, and Daan Wierstra.
\newblock Variational intrinsic control.
\newblock \emph{arXiv preprint arXiv:1611.07507}, 2016.

\bibitem[Haarnoja et~al.(2018{\natexlab{a}})Haarnoja, Hartikainen, Abbeel, and
  Levine]{haarnoja2018latent}
Tuomas Haarnoja, Kristian Hartikainen, Pieter Abbeel, and Sergey Levine.
\newblock Latent space policies for hierarchical reinforcement learning.
\newblock In \emph{International Conference on Machine Learning (ICML)},
  2018{\natexlab{a}}.

\bibitem[Haarnoja et~al.(2018{\natexlab{b}})Haarnoja, Zhou, Abbeel, and
  Levine]{haarnoja2018soft}
Tuomas Haarnoja, Aurick Zhou, Pieter Abbeel, and Sergey Levine.
\newblock Soft actor-critic: Off-policy maximum entropy deep reinforcement
  learning with a stochastic actor.
\newblock In \emph{International Conference on Machine Learning (ICML)},
  2018{\natexlab{b}}.

\bibitem[Harb et~al.(2018)Harb, Bacon, Klissarov, and Precup]{harb2018waiting}
Jean Harb, Pierre-Luc Bacon, Martin Klissarov, and Doina Precup.
\newblock When waiting is not an option: Learning options with a deliberation
  cost.
\newblock In \emph{Thirty-Second AAAI Conference on Artificial Intelligence},
  2018.

\bibitem[Harutyunyan et~al.(2019)Harutyunyan, Dabney, Borsa, Heess, Munos, and
  Precup]{harutyunyan2019termination}
Anna Harutyunyan, Will Dabney, Diana Borsa, Nicolas Heess, Remi Munos, and
  Doina Precup.
\newblock The termination critic.
\newblock \emph{arXiv preprint arXiv:1902.09996}, 2019.

\bibitem[Hausman et~al.(2018)Hausman, Springenberg, Wang, Heess, and
  Riedmiller]{hausman2018learning}
Karol Hausman, Jost~Tobias Springenberg, Ziyu Wang, Nicolas Heess, and Martin
  Riedmiller.
\newblock Learning an embedding space for transferable robot skills.
\newblock In \emph{International Conference on Learning Representations}, 2018.

\bibitem[Jinnai et~al.(2019)Jinnai, Park, Machado, and
  Konidaris]{jinnai2019exploration}
Yuu Jinnai, Jee~Won Park, Marlos~C Machado, and George Konidaris.
\newblock Exploration in reinforcement learning with deep covering options.
\newblock In \emph{International Conference on Learning Representations}, 2019.

\bibitem[Kamat \& Precup(2020)Kamat and Precup]{kamat2020diversity}
Anand Kamat and Doina Precup.
\newblock Diversity-enriched option-critic.
\newblock \emph{arXiv preprint arXiv:2011.02565}, 2020.

\bibitem[Khetarpal et~al.(2020)Khetarpal, Riemer, Rish, and
  Precup]{khetarpal2020towards}
Khimya Khetarpal, Matthew Riemer, Irina Rish, and Doina Precup.
\newblock Towards continual reinforcement learning: A review and perspectives.
\newblock \emph{arXiv preprint arXiv:2012.13490}, 2020.

\bibitem[Kirkpatrick et~al.(2017)Kirkpatrick, Pascanu, Rabinowitz, Veness,
  Desjardins, Rusu, Milan, Quan, Ramalho, Grabska-Barwinska,
  et~al.]{kirkpatrick2017overcoming}
James Kirkpatrick, Razvan Pascanu, Neil Rabinowitz, Joel Veness, Guillaume
  Desjardins, Andrei~A Rusu, Kieran Milan, John Quan, Tiago Ramalho, Agnieszka
  Grabska-Barwinska, et~al.
\newblock Overcoming catastrophic forgetting in neural networks.
\newblock \emph{Proceedings of the national academy of sciences}, 114\penalty0
  (13):\penalty0 3521--3526, 2017.

\bibitem[Klissarov \& Precup(2021)Klissarov and Precup]{klissarov2021flexible}
Martin Klissarov and Doina Precup.
\newblock Flexible option learning.
\newblock \emph{Advances in Neural Information Processing Systems}, 34, 2021.

\bibitem[Levy et~al.(2017)Levy, Konidaris, Platt, and Saenko]{levy2017learning}
Andrew Levy, George Konidaris, Robert Platt, and Kate Saenko.
\newblock Learning multi-level hierarchies with hindsight.
\newblock \emph{arXiv preprint arXiv:1712.00948}, 2017.

\bibitem[Lillicrap et~al.(2015)Lillicrap, Hunt, Pritzel, Heess, Erez, Tassa,
  Silver, and Wierstra]{lillicrap2015continuous}
T.~P. Lillicrap, J.~J. Hunt, A.~Pritzel, N.~Heess, T.~Erez, Y.~Tassa,
  D.~Silver, and D.~Wierstra.
\newblock Continuous control with deep reinforcement learning.
\newblock \emph{arXiv preprint arXiv:1509.02971}, 2015.

\bibitem[Machado et~al.(2017)Machado, Bellemare, and
  Bowling]{machado2017laplacian}
Marlos~C Machado, Marc~G Bellemare, and Michael Bowling.
\newblock A laplacian framework for option discovery in reinforcement learning.
\newblock In \emph{International Conference on Machine Learning}, pp.\
  2295--2304. PMLR, 2017.

\bibitem[McGovern \& Barto(2001)McGovern and Barto]{mcgovern2001automatic}
Amy McGovern and Andrew~G Barto.
\newblock Automatic discovery of subgoals in reinforcement learning using
  diverse density.
\newblock 2001.

\bibitem[Merel et~al.(2018)Merel, Hasenclever, Galashov, Ahuja, Pham, Wayne,
  Teh, and Heess]{merel2018neural}
Josh Merel, Leonard Hasenclever, Alexandre Galashov, Arun Ahuja, Vu~Pham, Greg
  Wayne, Yee~Whye Teh, and Nicolas Heess.
\newblock Neural probabilistic motor primitives for humanoid control.
\newblock \emph{arXiv preprint arXiv:1811.11711}, 2018.

\bibitem[Merel et~al.(2020)Merel, Tunyasuvunakool, Ahuja, Tassa, Hasenclever,
  Pham, Erez, Wayne, and Heess]{merel2020catch}
Josh Merel, Saran Tunyasuvunakool, Arun Ahuja, Yuval Tassa, Leonard
  Hasenclever, Vu~Pham, Tom Erez, Greg Wayne, and Nicolas Heess.
\newblock Catch \& carry: reusable neural controllers for vision-guided
  whole-body tasks.
\newblock \emph{ACM Transactions on Graphics (TOG)}, 39\penalty0 (4):\penalty0
  39--1, 2020.

\bibitem[Mnih et~al.(2015)Mnih, Kavukcuoglu, Silver, Rusu, Veness, Bellemare,
  Graves, Riedmiller, Fidjeland, Ostrovski, et~al.]{mnih2015human}
Volodymyr Mnih, Koray Kavukcuoglu, David Silver, Andrei~A Rusu, Joel Veness,
  Marc~G Bellemare, Alex Graves, Martin Riedmiller, Andreas~K Fidjeland, Georg
  Ostrovski, et~al.
\newblock Human-level control through deep reinforcement learning.
\newblock \emph{nature}, 518\penalty0 (7540):\penalty0 529--533, 2015.

\bibitem[Mo et~al.(2018)Mo, Li, Lin, and Lee]{mo2018adobeindoornav}
Kaichun Mo, Haoxiang Li, Zhe Lin, and Joon-Young Lee.
\newblock The adobeindoornav dataset: Towards deep reinforcement learning based
  real-world indoor robot visual navigation.
\newblock \emph{arXiv preprint arXiv:1802.08824}, 2018.

\bibitem[Nachum et~al.(2018)Nachum, Gu, Lee, and Levine]{nachum2018data}
Ofir Nachum, Shixiang Gu, Honglak Lee, and Sergey Levine.
\newblock Data-efficient hierarchical reinforcement learning.
\newblock \emph{arXiv preprint arXiv:1805.08296}, 2018.

\bibitem[Neunert et~al.(2020)Neunert, Abdolmaleki, Wulfmeier, Lampe,
  Springenberg, Hafner, Romano, Buchli, Heess, and
  Riedmiller]{neunert2020continuous}
Michael Neunert, Abbas Abdolmaleki, Markus Wulfmeier, Thomas Lampe, Tobias
  Springenberg, Roland Hafner, Francesco Romano, Jonas Buchli, Nicolas Heess,
  and Martin Riedmiller.
\newblock Continuous-discrete reinforcement learning for hybrid control in
  robotics.
\newblock In \emph{Conference on Robot Learning}, pp.\  735--751. PMLR, 2020.

\bibitem[Parisi et~al.(2019)Parisi, Kemker, Part, Kanan, and
  Wermter]{parisi2019continual}
German~I Parisi, Ronald Kemker, Jose~L Part, Christopher Kanan, and Stefan
  Wermter.
\newblock Continual lifelong learning with neural networks: A review.
\newblock \emph{Neural Networks}, 113:\penalty0 54--71, 2019.

\bibitem[Peng et~al.(2019)Peng, Kumar, Zhang, and Levine]{peng2019advantage}
Xue~Bin Peng, Aviral Kumar, Grace Zhang, and Sergey Levine.
\newblock Advantage-weighted regression: Simple and scalable off-policy
  reinforcement learning.
\newblock \emph{arXiv preprint arXiv:1910.00177}, 2019.

\bibitem[Pertsch et~al.(2020)Pertsch, Lee, and Lim]{pertsch2020accelerating}
Karl Pertsch, Youngwoon Lee, and Joseph~J Lim.
\newblock Accelerating reinforcement learning with learned skill priors.
\newblock \emph{arXiv preprint arXiv:2010.11944}, 2020.

\bibitem[Ramesh et~al.(2019)Ramesh, Tomar, and Ravindran]{ramesh2019successor}
Rahul Ramesh, Manan Tomar, and Balaraman Ravindran.
\newblock Successor options: An option discovery framework for reinforcement
  learning.
\newblock \emph{arXiv preprint arXiv:1905.05731}, 2019.

\bibitem[Riemer et~al.(2018)Riemer, Liu, and Tesauro]{riemer2018learning}
Matthew Riemer, Miao Liu, and Gerald Tesauro.
\newblock Learning abstract options.
\newblock \emph{arXiv preprint arXiv:1810.11583}, 2018.

\bibitem[Salter et~al.(2022)Salter, Hartikainen, Goodwin, and
  Posner]{salter2022priors}
Sasha Salter, Kristian Hartikainen, Walter Goodwin, and Ingmar Posner.
\newblock Priors, hierarchy, and information asymmetry for skill transfer in
  reinforcement learning.
\newblock \emph{arXiv preprint arXiv:2201.08115}, 2022.

\bibitem[Schulman et~al.(2017)Schulman, Wolski, Dhariwal, Radford, and
  Klimov]{schulman2017proximal}
J.~Schulman, F.~Wolski, P.~Dhariwal, A.~Radford, and O.~Klimov.
\newblock Proximal policy optimization algorithms.
\newblock \emph{arXiv preprint arXiv:1707.06347}, 2017.

\bibitem[Shiarlis et~al.(2018)Shiarlis, Wulfmeier, Salter, Whiteson, and
  Posner]{shiarlis2018taco}
Kyriacos Shiarlis, Markus Wulfmeier, Sasha Salter, Shimon Whiteson, and Ingmar
  Posner.
\newblock Taco: Learning task decomposition via temporal alignment for control.
\newblock \emph{arXiv preprint arXiv:1803.01840}, 2018.

\bibitem[Silver et~al.(2017)Silver, Schrittwieser, Simonyan, Antonoglou, Huang,
  Guez, Hubert, Baker, Lai, Bolton, et~al.]{silver2017mastering}
David Silver, Julian Schrittwieser, Karen Simonyan, Ioannis Antonoglou, Aja
  Huang, Arthur Guez, Thomas Hubert, Lucas Baker, Matthew Lai, Adrian Bolton,
  et~al.
\newblock Mastering the game of go without human knowledge.
\newblock \emph{nature}, 550\penalty0 (7676):\penalty0 354--359, 2017.

\bibitem[{\c{S}}im{\c{s}}ek \& Barto(2004){\c{S}}im{\c{s}}ek and
  Barto]{csimcsek2004using}
{\"O}zg{\"u}r {\c{S}}im{\c{s}}ek and Andrew~G Barto.
\newblock Using relative novelty to identify useful temporal abstractions in
  reinforcement learning.
\newblock In \emph{Proceedings of the twenty-first international conference on
  Machine learning}, pp.\ ~95, 2004.

\bibitem[Singh et~al.(2020)Singh, Liu, Zhou, Yu, Rhinehart, and
  Levine]{singh2020parrot}
Avi Singh, Huihan Liu, Gaoyue Zhou, Albert Yu, Nicholas Rhinehart, and Sergey
  Levine.
\newblock Parrot: Data-driven behavioral priors for reinforcement learning.
\newblock \emph{arXiv preprint arXiv:2011.10024}, 2020.

\bibitem[Smith et~al.(2018)Smith, {Van Hoof}, and Pineau]{Smith2019}
Matthew~J.A. Smith, Herke {Van Hoof}, and Joelle Pineau.
\newblock {An inference-based policy gradient method for learning options}.
\newblock In \emph{35th International Conference on Machine Learning, ICML
  2018}, volume~11, pp.\  7481--7490, 2018.
\newblock ISBN 9781510867963.
\newblock URL
  \url{https://pdfs.semanticscholar.org/809f/951c77b5a39e2a9d556e9cf9938de87f2393.pdf?{\_}ga=2.168186657.1832697831.1553020853-1357972575.1551696370}.

\bibitem[Solway et~al.(2014)Solway, Diuk, C{\'o}rdova, Yee, Barto, Niv, and
  Botvinick]{solway2014optimal}
Alec Solway, Carlos Diuk, Natalia C{\'o}rdova, Debbie Yee, Andrew~G Barto, Yael
  Niv, and Matthew~M Botvinick.
\newblock Optimal behavioral hierarchy.
\newblock \emph{PLoS computational biology}, 10\penalty0 (8):\penalty0
  e1003779, 2014.

\bibitem[Stolle \& Precup(2002)Stolle and Precup]{stolle2002learning}
Martin Stolle and Doina Precup.
\newblock Learning options in reinforcement learning.
\newblock In \emph{International Symposium on abstraction, reformulation, and
  approximation}, pp.\  212--223. Springer, 2002.

\bibitem[Sutton(1988)]{sutton1988learning}
Richard~S Sutton.
\newblock Learning to predict by the methods of temporal differences.
\newblock \emph{Machine learning}, 3\penalty0 (1):\penalty0 9--44, 1988.

\bibitem[Sutton et~al.(1999)Sutton, Precup, and Singh]{sutton1999between}
Richard~S Sutton, Doina Precup, and Satinder Singh.
\newblock Between mdps and semi-mdps: A framework for temporal abstraction in
  reinforcement learning.
\newblock \emph{Artificial intelligence}, 112\penalty0 (1-2):\penalty0
  181--211, 1999.

\bibitem[Wang et~al.(2019)Wang, Hu, and Scherer]{wang2019learning}
Wenshan Wang, Yaoyu Hu, and Sebastian Scherer.
\newblock Learning temporal abstraction with information-theoretic constraints
  for hierarchical reinforcement learning.
\newblock 2019.

\bibitem[Wulfmeier et~al.(2019)Wulfmeier, Abdolmaleki, Hafner, Springenberg,
  Neunert, Hertweck, Lampe, Siegel, Heess, and Riedmiller]{Wulfmeier2019}
Markus Wulfmeier, Abbas Abdolmaleki, Roland Hafner, Jost~Tobias Springenberg,
  Michael Neunert, Tim Hertweck, Thomas Lampe, Noah Siegel, Nicolas Heess, and
  Martin Riedmiller.
\newblock {Compositional Transfer in Hierarchical Reinforcement Learning}.
\newblock \penalty0 (1), 2019.
\newblock URL \url{http://arxiv.org/abs/1906.11228}.

\bibitem[Wulfmeier et~al.(2020)Wulfmeier, Rao, Hafner, Lampe, Abdolmaleki,
  Hertweck, Neunert, Tirumala, Siegel, Heess, et~al.]{wulfmeier2020data}
Markus Wulfmeier, Dushyant Rao, Roland Hafner, Thomas Lampe, Abbas Abdolmaleki,
  Tim Hertweck, Michael Neunert, Dhruva Tirumala, Noah Siegel, Nicolas Heess,
  et~al.
\newblock Data-efficient hindsight off-policy option learning.
\newblock \emph{arXiv preprint arXiv:2007.15588}, 2020.

\bibitem[Zhang \& Whiteson(2019)Zhang and Whiteson]{Zhang2019DAC}
Shangtong Zhang and Shimon Whiteson.
\newblock {DAC: The Double Actor-Critic Architecture for Learning Options}.
\newblock \penalty0 (NeurIPS), 2019.
\newblock URL \url{http://arxiv.org/abs/1904.12691}.

\end{thebibliography}
